\documentclass{article}

\usepackage{microtype}
\usepackage{graphicx}
\usepackage{subfigure}
\usepackage{booktabs} 

\usepackage{diagbox}
\usepackage{multirow}
\usepackage{colortbl}
\usepackage{makecell}

\usepackage{tabularray}
\UseTblrLibrary{diagbox}
\usepackage{amsmath}
\usepackage{upgreek}

\usepackage{hyperref}

\usepackage[accepted]{icml2024}

\usepackage{amsmath}
\usepackage{amssymb}
\usepackage{mathtools}
\usepackage{amsthm}

\usepackage[capitalize,noabbrev]{cleveref}

\theoremstyle{plain}
\newtheorem{theorem}{Theorem}[section]
\newtheorem{proposition}[theorem]{Proposition}
\newtheorem{lemma}[theorem]{Lemma}

\theoremstyle{definition}

\newtheorem{assumption}[theorem]{Assumption}
\newtheorem{notation}[theorem]{Notation}
\theoremstyle{remark}

\usepackage[textsize=tiny]{todonotes}

\usepackage[normalem]{ulem}

\icmltitlerunning{CRoFT: Robust Fine-Tuning with Concurrent Optimization for OOD Generalization and Open-Set OOD Detection}

\begin{document}

\twocolumn[
\icmltitle{CRoFT: Robust Fine-Tuning with Concurrent Optimization 
for\\ OOD Generalization and Open-Set OOD Detection}




\begin{icmlauthorlist}
\icmlauthor{Lin Zhu}{yyy}
\icmlauthor{Yifeng Yang}{yyy}
\icmlauthor{Qinying Gu}{comp}
\icmlauthor{Xinbing Wang}{yyy}
\icmlauthor{Chenghu Zhou}{yyy}
\icmlauthor{Nanyang Ye}{yyy}
\end{icmlauthorlist}

\icmlaffiliation{yyy}{Shanghai Jiao Tong University, Shanghai, China.}
\icmlaffiliation{comp}{Shanghai Artificial Intelligence Laboratory, Shanghai, China.}

\icmlcorrespondingauthor{Nanyang Ye}{ynylincoln@sjtu.edu.cn}

\icmlkeywords{Machine Learning, ICML}

\vskip 0.3in
]



\printAffiliationsAndNotice{} 

\begin{abstract}
Recent vision-language pre-trained models (VL-PTMs) have shown remarkable success in open-vocabulary tasks. However, downstream use cases often involve further fine-tuning of VL-PTMs, which may distort their general knowledge and impair their ability to handle distribution shifts. In real-world scenarios, machine learning systems inevitably encounter both covariate shifts (e.g., changes in image styles) and semantic shifts (e.g., test-time unseen classes). This highlights the importance of enhancing out-of-distribution (OOD) generalization on covariate shifts and simultaneously detecting semantic-shifted unseen classes. 
Thus a critical but underexplored question arises: \textit{How to improve VL-PTMs' generalization ability to closed-set OOD data, while effectively detecting open-set unseen classes during fine-tuning?} 
In this paper, we propose a novel objective function of OOD detection that also serves to improve OOD generalization. We show that minimizing the gradient magnitude of energy scores on training data leads to domain-consistent Hessians of classification loss,
a strong indicator for OOD generalization revealed by theoretical analysis.
Based on this finding, we have developed a unified fine-tuning framework that allows for concurrent optimization of both tasks. Extensive experiments have demonstrated the superiority of our method. The code is available at \url{https://github.com/LinLLLL/CRoFT}.
\vskip-0.2in
\end{abstract}



\section{Introduction}
Recent advances in large-scale vision-language pre-trained models (VL-PTMs), such as CLIP \cite{CLIP}, Grounding DINO \cite{liu2023grounding}, MiniGPT-4 \cite{zhu2023minigpt4}, etc., 
have shown promising results in visual-semantic learning.
However, in real-world scenarios, machine learning models often face challenges related to out-of-distribution (OOD) data, arising from disparities in data distributions between the training and test sets \cite{Meinshausen_2015, koh2020wilds}.
To address this issue, various paradigms \cite{zhou2021learning, wortsman2022model, andreassen2021evolution, li2021improved, mao2023context, jiang2023correlation, goyal2023finetune} have been proposed to fine-tune VL-PTMs, aiming to enhance their robustness against test-time distribution shifts. 
By adopting these fine-tuning techniques, the fine-tuned VL-PTMs can quickly adapt to downstream tasks even with only a few training examples \cite{fan2021generalized, nakamura2019revisiting, gao2020making}.
\begin{figure}
\centering
\centerline{\includegraphics[width=0.98\linewidth]{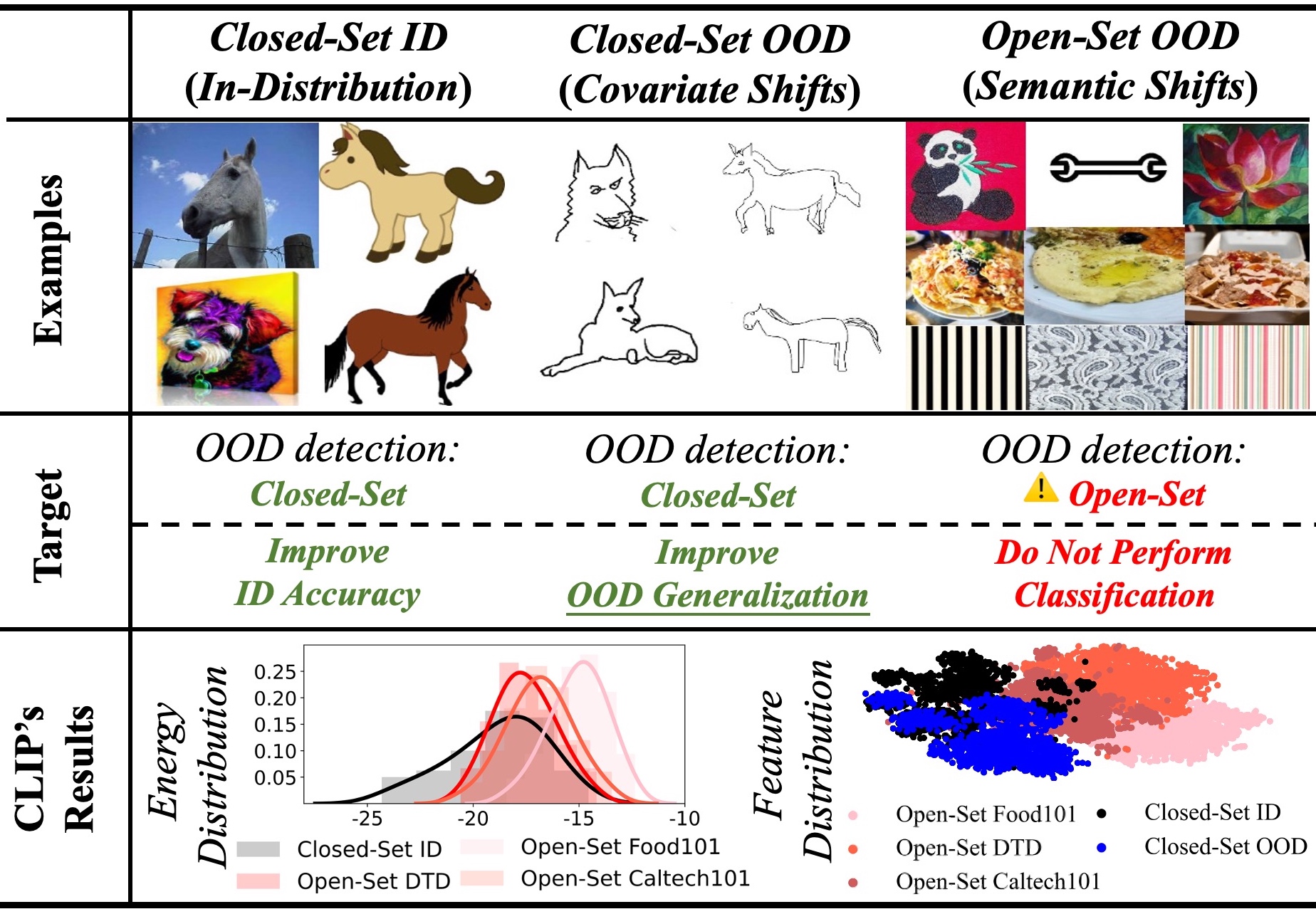}}
\vskip -0.1in
\caption{Illustration of typical data setting in real-world scenarios.
For example, we may encounter various types of data in real-world applications: (i) closed-set ID data (e.g., dog), (ii) closed-set OOD data with covariate shifts (e.g., dog with changed image styles), and (iii) open-set OOD data with semantic shifts (e.g., panda).
\textbf{
The significant overlaps in energy distributions between closed-set ID and open-set OOD data pose a challenge for CLIP in detecting open-set OOD data. The notable discrepancy between the closed-set ID and closed-set OOD data also complicates achieving OOD generalization for closed-set OOD data.} 
}
\label{fig:setups}
\vskip -0.2in
\end{figure}
Furthermore, some methods have shown improved generalization performances within the same downstream tasks under OOD scenarios \cite{wortsman2022model, goyal2023finetune, zhu2023prompt}. 
Nevertheless, most of these methods primarily focused on closed-set visual concepts, limiting the model to a pre-defined list of categories. Their discrimination ability for new categories unseen during training has not been thoroughly explored.

Unfortunately, as illustrated in Figure~\ref{fig:setups}, real-world testing environments often involve known classes living in diverse environments, as well as new categories unseen during training \cite{wang2023towards, bai2023feed}. It is crucial to distinguish these unknown categories from known ones, rather than blindly predicting them as known classes. The ability to detect unseen classes is particularly essential for ensuring system safety in high-risk applications, such as autonomous driving \cite{liu2021visual, majee2021few} and medical imaging \cite{castro2020causality}. 

Previous studies have focused on improving models' robustness when fine-tuning VL-PTMs or developing methods for unseen-class recognition independently. Consequently, existing approaches are highly specialized in one task, but not capable of handling both aspects simultaneously. This raises a critical but underexplored question:

\textbf{\textit{When fine-tuning VL-PTMs to downstream tasks, how to improve models' generalization ability to closed-set OOD data, while effectively detecting open-set unseen classes during fine-tuning?}}

In this paper, we develop a novel fine-tuning paradigm to go beyond the limitations of previous studies that were unable to address both aspects simultaneously. Initially, leveraging the widely used energy-based function \cite{liu2020energy} for detecting unknown classes, we propose an energy distribution reshaping (EDR) loss. The proposed EDR loss aims to approach an optimal solution of minimizing energy scores on in-distribution (ID) data, which is implemented by minimizing the gradient magnitude of energy scores.
This enables us to fine-tune VL-PTMs in the direction of distinguishing the energy distribution of known classes from other distributions.

Furthermore, by connecting the two challenges using Hessians, we show that the proposed EDR loss theoretically leads to domain-consistent Hessians, thereby helping to bound the generalization performance on closed-set OOD test data \cite{rame2022fishr, hemati2023understanding}.
Building upon this finding, we have developed a novel Hessian-based OOD generalization bound, which is associated with model performance under worst-case OOD scenarios. Motivated by bound minimization, we introduce a unified fine-tuning framework named \textbf{CRoFT}. This framework is designed to achieve \textit{\textbf{ro}bust \textbf{f}ine-\textbf{t}uning while enabling \textbf{c}oncurrent optimization for both aspects}.
Through the use of different data settings to evaluate model performance, we have demonstrated that our CRoFT approach can obtain state-of-the-art results on both tasks, especially showcasing up to 20\% improvements in detecting open-set unseen classes.



\section{Problem Setting}
\label{sec:problem_setting}
\textbf{Task definition}~Based on the \textit{closed-set ID} samples, $\{\mathbf{x^{(i)}},\mathbf{y^{(i)}}\}_{i=1}^N$, sampled from some source domain $\mathcal{S}$, 
our goal is to fine-tune a VL-PTM to obtain a robust predictor $f: \mathcal{X} \rightarrow \mathcal{Y}$, which maps inputs $\mathbf{x} \in \mathcal{X}=\mathbb{R}^{D}$ to outputs $\mathbf{y} \in \mathcal{Y}=\mathbb{R}^K$ (where $K$ is the class number and $D$ is the dimension of $\mathbf{x}$).  
In open-set scenarios, different distribution shifts can occur. These scenarios involve \textit{closed-set OOD} data that exhibit \textit{covariate shifts} (i.e., changes in environments, while class labels remain the same as the ID data) and \textit{open-set OOD} data with \textit{semantic shifts} (i.e., test-time unseen classes). 
Therefore, our focus lies in enhancing the robustness of predictor $f$ from two perspectives: 1) \textit{OOD generalization}, which is related to testing the model on known classes from a new domain, ${\mathcal{T}}$, that exhibit covariate shifts; 2) \textit{open-set OOD detection}, enabling the fine-tuned model to detect test-time unseen classes.

Motivated by the remarkable success of the large-scale
pre-trained vision-language model CLIP \cite{CLIP} in learning general visual knowledge, we delve into exploring a CLIP-based framework for boosting both OOD generalization and open-set OOD detection across diverse data scenarios. It's important to note that our approach readily extends to other VL-PTMs, such as ALIGN \cite{li2021align}, BLIP-2 \cite{li2023blip2}, Grounding DINO \cite{liu2023grounding}, and MiniGPT-4 \cite{zhu2023minigpt4}, by employing a contrastive loss technique to align image representations with the corresponding text representations \cite{li2022supervision, goel2022cyclip, mu2022slip}.

\textbf{First look at CLIP's performance}
Based on the real-world data setting depicted in Figure~\ref{fig:setups}, we initially assess CLIP's performance on the two challenging tasks. Details of the data setting are in Setup-II of \cref{sec:exp}.
Utilizing the widely adopted energy score \cite{liu2020energy} for detecting unseen classes, we visualize the energy distribution of different types of data.
As illustrated in Figure~\ref{fig:setups}, there is a significant overlap between the energy distributions of closed-set and open-set samples. This overlap poses a challenge for the CLIP model to effectively distinguish between them, making it difficult to detect open-set OOD instances. 
Furthermore, we employ t-SNE \cite{van2008visualizing} to reduce CLIP's image features to a 2-dimensional space and provide the embedding visualization in Figure~\ref{fig:setups}. It is shown that there is a noticeable discrepancy between the feature distributions of closed-set ID data and closed-set OOD data, which complicates the task of achieving OOD generalization for closed-set OOD datasets. 


Therefore, addressing the problem of enabling VL-PTMs to handle various distribution shits, especially enhancing VL-PTMs' ability to detect open-set OOD data, is a key research direction that requires urgent attention.
Before delving into our approach, we provide an explanation for the model assumption as described in Assumption~\ref{ass:xfinite}.
\begin{assumption}
 We use notation $\mathcal{D}$ to represent a distribution on input space $\mathcal{X}$. 
    Given an empirical distribution (denoted as $\widehat{\mathcal{D}}_\mathcal{S}=\{\mathbf{x^{(i)}}\}_{i=1}^N$) sampled from the input space of source domain $\mathcal{S}$, we input images and closed-set class names into the CLIP model. We then obtain the zero-shot image features and text features, denoted as $\{\mathbf{z^{(i)}_{I0}}\}_{i=1}^N$ and $\{\mathbf{z^{(i)}_{T0}}\}_{i=1}^K$, respectively. 
    In the fine-tuning framework, we represent the fine-tuned image features and text features as $\mathbf{z_{I}^{(i)}}:=\mathbf{z_{I}}(\mathbf{x^{(i)}};\boldsymbol{\uptheta})~(i=1,\cdots,N)$ and $\mathbf{z^{(j)}_{T}}:=\mathbf{z^{(j)}_{T}}(\boldsymbol{\uptheta})~(j=1,\cdots,K)$, respectively.
    $\boldsymbol{\uptheta}$ denotes the model parameter. 
    Consider an bounded instance loss function $\ell$ such that $\mathcal{Y}\times\mathcal{Y} \rightarrow[0,c]$, and $\ell(\mathbf{y_1}, \mathbf{y_2})=0$ if and only if $\mathbf{y_1}=\mathbf{y_2}$ ($\mathbf{y_1} \in \mathcal{Y}, \mathbf{y_2} \in \mathcal{Y}$).
    The expected risk on domain $\mathcal{D}$ is represented as $\mathcal{E}_{\mathcal{D}}(\boldsymbol{\uptheta}) := \mathbb{E}_\mathcal{D}(f(\mathbf{x};\mathbf{\boldsymbol{\uptheta}});\mathbf{y})$.  
    \label{ass:xfinite}
\end{assumption}

\section{Methodology}
\subsection{Reshaping energy distribution for open-set OOD detection}
Despite various fine-tuning strategies proposed to improve robustness in classifying closed-set data, their limitations 
in detecting the open-set unseen classes
are largely overlooked. Motivated by the promising results of the CLIP model in learning general visual-semantic knowledge, we propose a unified fine-tuning framework to improve CLIP's OOD generalization ability while enabling the model to detect open-set unseen classes. Based on the widely used energy function \cite{liu2020energy} for OOD detection, we reshape the energy distribution of the fine-tuned model's output, which is expected to facilitate discrimination between closed-set samples and open-set samples. 

The energy score \cite{liu2020energy}, as commonly used for OOD detection, is trained to assign lower energy values to more plausible or confident configurations. 
In the regime of fine-tuning CLIP, the energy score is calculated as:
\begin{equation}
    E_{\boldsymbol{\uptheta}}(\mathbf{x}) = -\mathbb{E}_{\mathbf{x}\in \mathcal{X}}\log\sum_{i=1}^K\exp\left<\mathbf{z_I}(\mathbf{x};\boldsymbol{\uptheta}), \mathbf{z_T^{(i)}}(\boldsymbol{\uptheta})\right>
\label{eq:energy_score}
\end{equation}
Minimizing energy scores on training data has been demonstrated as an effective approach in previous studies \cite{du2019implicit, katz2022training}. In this paper, we focus on the optimization objective of $\min E_{\boldsymbol{\uptheta}}(\mathbf{x})$. With this training criterion, we define the empirical loss $\mathcal{L}(\mathcal{D}_\mathcal{S}, \boldsymbol{\uptheta})=-\frac1N\sum_{i=1}^N E_{\boldsymbol{\uptheta}}(\mathbf{x^{(i)}})$ with the training datasets $\mathcal{D}_\mathcal{S}=\{\mathbf{x^{(i)}}\}_{i=1}^N$. 
By utilizing the gradient decent \cite{amari1993backpropagation} method, model parameter $\boldsymbol{\uptheta}$ can be updated with the following gradient:
\vskip-0.16in
\begin{equation}
    \frac{\partial\mathcal{L}(\mathcal{D}, \boldsymbol{\uptheta})}{\partial \boldsymbol{\uptheta}} 
    =\frac1N\sum_{i=1}^N\frac{\partial E_{\boldsymbol{\uptheta}}(\mathbf{x^{(i)}})}{\partial{\boldsymbol{\uptheta}}}
\label{eq:gradient}
\end{equation}

\vskip-0.12in
For large-size VL-PTMs, calculating the gradient as defined in Equation~\ref{eq:gradient} along with the whole model is computationally expensive. To address this issue, we adopt lightweight fine-tuning for training efficiency. Specifically, we inject one-layer linear projections into CLIP's image encoder and text encoder, respectively, while keeping parameters (${\boldsymbol{\uptheta}}_0$) in pre-trained encoders frozen. By doing this, we only need to calculate the gradient of loss $\mathcal{L}(\mathcal{D}_\mathcal{S}, \boldsymbol{\uptheta})$ with respect to parameters in linear projections (${\boldsymbol{\uptheta}}_l$). 
Finally, we can approach the optimum in Equation~\ref{eq:gradient} 
as illustrated in Proposition~\ref{prop:loss}.

\begin{proposition}{\textbf{[Energy distribution reshaping (EDR) loss]}}
Given the training data $\widehat{\mathcal{D}}_\mathcal{S}$, in our fine-tuning framework, we calculate the training data's energy scores based on Equation~\ref{eq:energy_score}. To approach 
the solution of ${\min E_{\boldsymbol{\uptheta}}(\mathbf{x})}$, i.e., ${\nabla_{\boldsymbol{\uptheta}} E_{\boldsymbol{\uptheta}}(\mathbf{x})} \rightarrow \boldsymbol{0}$,
we propose to minimize the squared norm of Equation~\ref{eq:gradient} (i.e., magnitude of the gradient vector), which is formulated as optimizing the following loss:
\begin{equation}
\small{
\mathcal{L}_{\text{e}}=\frac1N\sum_{i=1}^N\left\|\nabla_{\boldsymbol{\uptheta_l}}\left(\log\sum_{j=1}^K\exp\left<\mathbf{{\mathbf{z_I}(\mathbf{x^{(i)}};\boldsymbol{\uptheta})}, z_T^{(j)}}(\boldsymbol{\uptheta})\right>\right)\right\|_2^2}
\label{eq:loss_edr}
\end{equation}
where ${\boldsymbol{\uptheta}} = \{\boldsymbol{\uptheta}_0, \boldsymbol{\uptheta}_l\}$, ${\boldsymbol{\uptheta}}_0$ is the frozen parameter in CLIP's pre-trained encoders, and ${\boldsymbol{\uptheta}}_l$ is the parameter in linear projections that need to be optimized.
\label{prop:loss}
\end{proposition}

\vskip-0.1in
Unlike previous studies \cite{liu2020energy, katz2022training, tonin2021unsupervised} that directly reduce energy scores of ID training data, \textbf{we find the proposed EDR loss not only enhances the open-set OOD detection abilities but is also secretly helping to improve OOD generalization, as demonstrated in Theorem~\ref{Th:2}.} 

\begin{figure*}
\centering
\centerline{\includegraphics[width=1.0\linewidth]{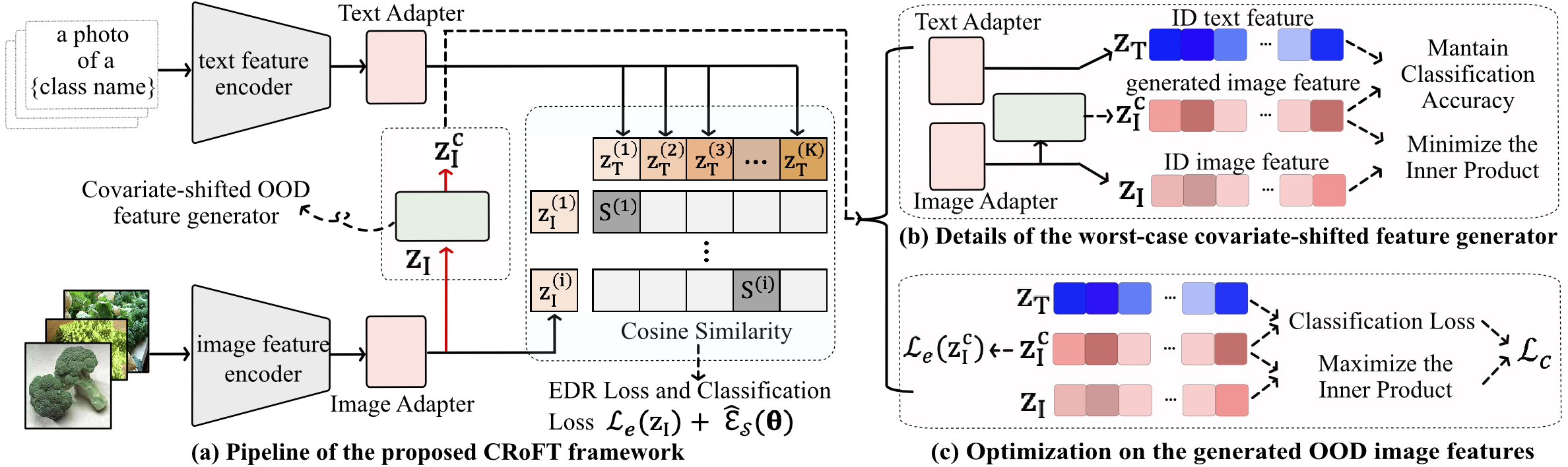}}
\vskip -0.1in
\caption{\textbf{Overview of our CRoFT framework}. 
Our theoretical analysis leads to the design of a new fine-tuning framework. As shown in Figure~(a), we inject adapters, i.e., one-layer linear projections after the CLIP's pre-trained encoders. 
Based on the adapted image feature $\mathbf{z_I}$ and adapted text feature $\mathbf{z_T}$, we generate the most challenging covariate-shifted OOD image features $\mathbf{z_I^c}$, simulating the worst-case scenarios.
The corresponding generation process, depicted in Figure~(b), follows the criterion defined in Equation~\ref{eq:ood_generator}, which preserves semantic information to maintain classification accuracy but differs from the ID image feature $\mathbf{z_I}$.
Finally, as shown in Figure~(c), we optimize on the generated $\mathbf{z_I^c}$ using the proposed loss $\mathcal{L}_{\text{c}}$. Meanwhile, we minimize the classification loss (cross-entropy) on the ID image features, denoted as $\widehat{\mathcal{E}}_{\mathcal{S}}({\boldsymbol{\uptheta}})$, 
while reshaping the energy distribution for $\mathbf{z_I}$ and  $\mathbf{z_I^c}$ through the EDR loss (i.e., $\mathcal{L}(\mathbf{z_I})$ and  $\mathcal{L}(\mathbf{z_I^c})$). 
}
\label{fig:overview}
\vskip-0.2in
\end{figure*}

\subsection{Minimizing the EDR-inspired OOD generalization bound for better concurrent optimization}
In this section, we start by developing an EDR-inspired OOD generalization bound under the worst-case scenarios, where the theoretical link between the EDR loss and the generalization bound is detailed in \cref{sec:connection}.
The novel bound indicates that the OOD generalization performance on target domain is related to the performance gap between the minimum risk on source domain and the empirical risk on worst-case covariate-shifted data. 
Motivated by reducing this performance gap, we introduce an adversarial-learning-based approach. Specifically, we iteratively generate the most challenging covariate-shifted image features, simulating the worst-case scenarios,
and enhances the model's robustness for these generated OOD image features.
Model assumptions in this section are presented in Assumption~\ref{ass:xfinite2}.

\begin{assumption}
Given empirical distributions generated with $m$ i.i.d. samples from the source domain and target domain, denoted as $\widehat{\mathcal{D}}_\mathcal{S}$ and $\widehat{\mathcal{D}}_\mathcal{T}$, respectively, we denote the empirical risk on $\widehat{\mathcal{D}}_\mathcal{S}$ ( $\widehat{\mathcal{D}}_\mathcal{T}$) as $\widehat{\mathcal{E}}_\mathcal{S}(\boldsymbol{\uptheta})$ ($\widehat{\mathcal{E}}_\mathcal{T}(\boldsymbol{\uptheta})$).
Let $\widehat{\mathcal{D}}_\mathcal{S}^c$ represent the empirical distribution of the worst-case covariate-shifted OOD data, while sharing the same semantic information with $\widehat{\mathcal{D}}_\mathcal{S}$.
We denote the empirical risk on the worst-case covariate-shifted OOD data as $\widehat{\mathcal{E}}^c_\mathcal{S}(\boldsymbol{\uptheta})$. 
Based on the distribution distance \cite{zhao2018adversarial, cha2021swad}: $\operatorname{Div}(\mathcal{D}_\mathcal{S},\mathcal{D}_\mathcal{T})=2{\sup}_A|{\Pr}_{\mathcal{D}_\mathcal{S}}(A)-{\Pr}_{\mathcal{D}_\mathcal{T}}(A)|$, we assume that there exists $\varepsilon_c>0$, such that $\operatorname{Div}(\widehat{\mathcal{D}}_\mathcal{S}^c,\widehat{\mathcal{D}}_\mathcal{T}) \leq \operatorname{Div}(\widehat{\mathcal{D}}_\mathcal{S},\widehat{\mathcal{D}}_\mathcal{T}) +\varepsilon_c$.
For each image feature ${\mathbf{z}}_\mathbf{I}$ from the source domain, 
it is assumed that the corresponding image features $\Tilde{\mathbf{z}}_\mathbf{I}$ from the target domain, which share the same label with ${\mathbf{z}}_\mathbf{I}$, satisfy the condition $\vert\vert \mathbf{z_{I}} - \Tilde{\mathbf{z}}_\mathbf{I} \vert\vert_2 \leq \varepsilon^\prime$, where $\varepsilon^\prime$ is a small positive value.
\label{ass:xfinite2}
\end{assumption}

\begin{theorem}\textbf{[Hessian-based generalization bound]}
Let the VC dimension of the parameter space $\Theta$ be $v$.
Let ${\mathbf{H}}_\mathcal{S}(\boldsymbol{\uptheta}_l)$ denote the Hessian matrix of the expected risk on the source domain with respect to $\boldsymbol{\uptheta}_l$.
Assume that the Hessian matrices of the risk functions on both the source and target domains satisfy the $\rho$-Lipschitz condition.
For any $\mathcal{D}\in\{\mathcal{S},\mathcal{T}\}$
and any local perturbation
$\boldsymbol{\vartheta}\in B_r(\mathbf{0})$,
suppose that the corresponding curvature term in Lemma~\ref{lemma:note_for_bound} satisfies
$\frac12 \left|\boldsymbol{\vartheta}^\top
\mathbf{C}_\mathcal{D}(\boldsymbol{\uptheta}^\ast)
\boldsymbol{\vartheta}
\right| \le \varepsilon_H$.
Then, for $0 \le \delta \le 1$, with probability at least $1-\delta$, we have:
\begin{equation}
\begin{aligned}
\mathcal{E}_\mathcal{T}(\boldsymbol{\uptheta})
&\le
\widehat{\mathcal{E}}_\mathcal{S}^c(\boldsymbol{\uptheta})
-\min_{\boldsymbol{\uptheta}'}
\widehat{\mathcal{E}}_\mathcal{S}(\boldsymbol{\uptheta}')
+\frac{1}{2}
\left|
\boldsymbol{\vartheta}^\top
{\mathbf{H}}_\mathcal{S}(\boldsymbol{\uptheta}^*)
\boldsymbol{\vartheta}
\right|
\\
&+\operatorname{Div}(\widehat{\mathcal{D}}_\mathcal{S},
\widehat{\mathcal{D}}_\mathcal{T})
+\max\{\lambda_\mathcal{T},
2\lambda_\mathcal{S}-\lambda_\mathcal{T}\}
+\lambda
\\
&+\varepsilon
+\varepsilon_H 
+\varepsilon_c
+O\left(
\sqrt{\frac{v\log(m/v)+\log(1/\delta)}{m}}
\right),
\end{aligned}
\end{equation}
where $\varepsilon=O(\varepsilon^\prime)
+\frac{\rho}{6}r^3$,
$\lambda=\mathcal{E}_\mathcal{S}(\boldsymbol{\uptheta}^*)
+\mathcal{E}_\mathcal{T}(\boldsymbol{\uptheta}^*)
$, $\varepsilon^\prime$ and $\varepsilon_c$ are small positive values as defined in Assumption~\ref{ass:xfinite2}, $\lambda =\mathcal{E}_{\mathcal{S}}(\boldsymbol{\uptheta}^*) + \mathcal{E}_{\mathcal{T}}(\boldsymbol{\uptheta}^*)$, 
$\lambda_\mathcal{S} =\mathcal{E}_\mathcal{S}(\boldsymbol{\uptheta}^*)$, $\lambda_\mathcal{T} =\mathcal{E}_\mathcal{T}(\boldsymbol{\uptheta}^*)$,
and $\boldsymbol{\uptheta}^*$ is the risk-minimizing optimum on the combined source and target data.
\label{Th:1}
\end{theorem}

This theorem indicates that the generalization performance on the target domain $\mathcal{T}$ is related to the performance gap between minimum risk on source domain, i.e, $\min_{\boldsymbol{\uptheta}^\prime}{\widehat{\mathcal{E}}}_\mathcal{S}({\boldsymbol{\uptheta}}^\prime)$, and empirical risk on the worst-case covariate-shifted OOD samples, i.e, $\widehat{\mathcal{E}}_\mathcal{S}^c({\boldsymbol{\uptheta}})$. Motivated by bound minimization, we thus aim to optimize the model performances on the worst-case covariate-shifted OOD scenarios, reducing performance degradation on the target domain. To achieve this, we propose to generate the worst-case covariate-shifted image feature $\mathbf{{z}_\mathbf{I}^c}$, which preserves semantic information for accurate classification but differs from the ID image feature $\mathbf{z_I}$. Formally, we formulate this procedure as below:

\begin{proposition}{\textbf{[Optimization on the worst-case covariate-shifted OOD features]}}
Utilizing a covariate-shifted OOD feature generator $g(\cdot)$, which conducts a one-layer linear projection on ID image features, we generate the worst-case covariate-shifted image features as:
\vskip-0.15in
\begin{equation}
\operatorname{argmin}_{g}
 \frac{\lambda_1}{N} \sum_{i=1}^N\left<g\left(\mathbf{z_I^{(i)}}\right), \mathbf{z_I^{(i)}}\right> + \widehat{\mathcal{E}}_{\mathcal{S}}^c(\boldsymbol{\uptheta})
 \label{eq:ood_generator}
\end{equation}
\vskip-0.15in
where $g(\mathbf{z_I^{(i)}})$ is the generated worst-case OOD image feature, and 
$\lambda_1$ is a hyperparameter to balance covariate shifts and classification accuracy. Then we 
optimize performance on the generated covariate-shifted features by minimizing:
\vskip-0.15in
\begin{equation}
   \mathcal{L}_{\text{c}} =-
   \frac{\lambda_1}{N} \sum_{i=1}^N\left<g\left(\mathbf{z_I^{(i)}}\right), \mathbf{z_I^{(i)}}\right>+
   \widehat{\mathcal{E}}_{\mathcal{S}}^c(\boldsymbol{\uptheta})
   \label{loss:coo}
\end{equation}
\vskip-0.15in
\label{prop:wccf}
\end{proposition}

\subsection{Theoretical connection between the EDR loss and OOD generalization}
\label{sec:connection}
In previous studies \cite{rame2022fishr, hemati2023understanding}, it has been discovered that matching simultaneously domain-level risks and Hessians can improve OOD generalization.
Different from previous works, we do not directly regularize the model to learn domain-consistent Hessians. Instead, we theoretically demonstrate that the proposed EDR loss for open-set OOD detection can secretly lead to domain-consistent Hessinas of classification loss, thereby enabling concurrent optimization for both OOD generalization and OOD detection.
Before delving into the details, we introduce an important property to show the relationship between the EDR loss and the Hessians, as illustrated in Lemma~\ref{lemma:note_for_bound}.


\begin{lemma}{\textbf{[Domain-consistent Hessian decomposition induced by the EDR loss]}}
\label{lemma:note_for_bound}
For each input $\mathbf {x^{(i)}}$, define the inner product between its image feature and the text feature of the ground-truth class as
\[
S^{(i)}=\left<\mathbf{z_I}(\mathbf{x^{(i)}};\boldsymbol{\uptheta}),
\mathbf{z_T^{(i)}}(\boldsymbol{\uptheta})\right>.
\]
Let
$\widehat{\mathbf H}_{\mathcal D}(\boldsymbol{\uptheta}_l)$
denote the Hessian matrix of the empirical risk
$\widehat{\mathcal E}_{\mathcal D}(\boldsymbol{\uptheta})$
with respect to the adapter parameters
$\boldsymbol{\uptheta}_l$.
Define
\begin{equation}
A^{(i)}(\boldsymbol{\uptheta})
:=-E_{\boldsymbol{\uptheta}}(\mathbf{x^{(i)}})=
\log
\sum_{j=1}^{K}
\exp
\left<
\mathbf{z_I}(\mathbf{x^{(i)}};\boldsymbol{\uptheta}),
\mathbf{z_T^{(j)}}(\boldsymbol{\uptheta})
\right>,
\end{equation}
and denote
\begin{equation}
\mathbf C_{\mathcal D}(\boldsymbol{\uptheta}_l)
:=
\frac1N
\sum_{i=1}^N
\nabla_{\boldsymbol{\uptheta}_l}^2
A^{(i)}(\boldsymbol{\uptheta}).
\end{equation}
\begin{equation}
\mathbf M_{\mathcal D}(\boldsymbol{\uptheta}_l)
:
=-\frac1N
\sum_{i=1}^N
\begin{bmatrix}
\mathbf 0 &
\mathbf {z_{I0}^{(i)}}
\mathbf {z_{T0}^{(i)\top}}
\\
\mathbf {z_{T0}^{(i)}}
\mathbf {z_{I0}^{(i)\top}}
&
\mathbf 0
\end{bmatrix}.
\end{equation}
Then the empirical Hessian admits the following decomposition:
\begin{equation}
\begin{aligned}
\widehat{\mathbf H}_{\mathcal D}(\boldsymbol{\uptheta}_l)
=
\mathbf C_{\mathcal D}(\boldsymbol{\uptheta}_l)
+\mathbf M_{\mathcal D}(\boldsymbol{\uptheta}_l).
\end{aligned}
\end{equation}

Furthermore, let
\[
g_i(\boldsymbol{\uptheta}_l)
=\nabla_{\boldsymbol{\uptheta}_l}
A^{(i)}(\boldsymbol{\uptheta}),
\qquad
\mathbf J_i(\boldsymbol{\uptheta}_l)
=\nabla_{\boldsymbol{\uptheta}_l}^2
A^{(i)}(\boldsymbol{\uptheta}).
\]
Then a local optimum $\boldsymbol{\uptheta}^\ast$ of the EDR loss satisfies
\begin{equation}
\sum_{i=1}^N
\mathbf J_i(\boldsymbol{\uptheta}^\ast)^\top
g_i(\boldsymbol{\uptheta}^\ast)
=
\mathbf 0.
\end{equation}
Therefore, the local optimality condition of the EDR loss constrains the action of
$\nabla_{\boldsymbol{\uptheta}_l}^2 A^{(i)}(\boldsymbol{\uptheta})$
along the gradient direction
$\nabla_{\boldsymbol{\uptheta}_l} A^{(i)}(\boldsymbol{\uptheta})$,
thereby controlling the curvature response of
$\mathbf C_{\mathcal D}(\boldsymbol{\uptheta}_l)$
in the gradient direction.

Moreover, let
$
B_r(\mathbf{0})
=\left\{\boldsymbol{\vartheta}\in
\mathbb{R}^d : \| \boldsymbol{\vartheta} \|_2 \le r \right\}$
denote the Euclidean closed ball centered at the origin $\mathbf{0}$ with radius $r$.
Within the local optimal region of the EDR loss, i.e., Eq.~\eqref{eq:loss_edr}, if the following curvature-control condition holds:
\begin{equation}
\frac12
\left|
\boldsymbol{\vartheta}^{\top}
\mathbf C_{\mathcal D}(\boldsymbol{\uptheta}^{\ast})
\boldsymbol{\vartheta}
\right|
\le
\varepsilon_H,
\qquad
\forall \boldsymbol{\vartheta}\in B_r(\mathbf 0),
\end{equation}
then the dominant term of the Hessian is determined by the domain-consistent second-order term
$\mathbf M_{\mathcal D}(\boldsymbol{\uptheta}_l)$.
\end{lemma}

\begin{theorem}{\textbf{[The EDR loss bound OOD generalization]}}
Let $\boldsymbol{\uptheta}^\ast$ be a common stationary point of the source-domain risk and the target-domain risk.
Assume that the following conditions hold in a local neighborhood
$B_r(\boldsymbol{\uptheta}^\ast)$:
(1) The Hessian matrices of both the source and target risks are $\rho$-Lipschitz continuous;
(2) For any $\mathcal{D}\in\{\mathcal{S},\mathcal{T}\}$ and any local perturbation
$\boldsymbol{\vartheta}\in B_r(\mathbf{0})$, the curvature term defined in Lemma~\ref{lemma:note_for_bound} satisfies
$\frac12 \left| \boldsymbol{\vartheta}^\top
\mathbf{C}_\mathcal{D}(\boldsymbol{\uptheta}^\ast)
\boldsymbol{\vartheta} \right| \le \varepsilon_H$
Let
$\varepsilon=O(\varepsilon^\prime)
+\frac{\rho}{6}r^3 $.
Then, for any local perturbation
$\boldsymbol{\vartheta}\in B_r(\mathbf{0})$, the following OoD generalization bound holds:
\begin{equation}
\begin{aligned}
\max_{\boldsymbol{\uptheta}\in B_r(\boldsymbol{\uptheta}^\ast)}
&\left|
\widehat{\mathcal E}_{\mathcal{T}}
(\boldsymbol{\uptheta})
-\widehat{\mathcal E}_{\mathcal{S}}
(\boldsymbol{\uptheta}^\ast)
\right|
\le \left|
\widehat{\mathcal E}_{\mathcal{T}}
(\boldsymbol{\uptheta}^\ast)
-\widehat{\mathcal E}_{\mathcal{S}}
(\boldsymbol{\uptheta}^\ast)\right|
\\
&+\frac12
\left|
\boldsymbol{\vartheta}^\top
{\mathbf{H}}_{\mathcal{S}}
(\boldsymbol{\uptheta}^\ast)
\boldsymbol{\vartheta}
\right|
+\varepsilon
+2\varepsilon_H .
\end{aligned}
\label{eq:Taylor_expansion_corrected}
\end{equation}
Here,
$B_r(\boldsymbol{\uptheta}^\ast)=
\left\{\boldsymbol{\uptheta}\in \mathbb{R}^d
:\| \boldsymbol{\uptheta}-\boldsymbol{\uptheta}^\ast \|_2 \le r
\right\}$
denotes the Euclidean closed ball centered at $\boldsymbol{\uptheta}^\ast$ with radius $r$.
\label{Th:2}
\end{theorem}
Therefore, by connecting the EDR loss with the Hessians of empirical classification loss, we theoretically discover that the EDR loss can lead to a bound of the performance gap between closed-set ID data and closed-set OOD data. This implies that optimizing for open-set OOD detection with EDR loss also involves optimizing for OOD generalization.

\subsection{Algorithm of the proposed CRoFT}
\label{sec:algorithm}
Our theoretical analysis thus leads to the design of a new fine-tuning framework with concurrent optimization for both tasks.
As illustrated in Figure~\ref{fig:overview}, we prioritize computational efficiency by adopting lightweight fine-tuning techniques.
We employ an image adapter and a text adapter after CLIP’s image encoder and text encoder, respectively, while keeping parameters in pre-trained encoders frozen. Both the image adapter and text adapter are implemented as one-layer linear projections. 
After that, we proceed with the adversarial learning procedure as discussed in Proposition~\ref{prop:wccf}, to generate covariate-shifted OOD image features. The generation process is illustrated in Figure~\ref{fig:overview}~(b).
Finally, we minimize the classification loss on the ID image features, denoted as $\widehat{\mathcal{E}}_{\mathcal{S}}({\boldsymbol{\uptheta}})$, while incorporating two regularization terms. 1) We optimize on the generated OOD image features $\mathbf{z_I^c}$ using the proposed $\mathcal{L}_{\text{c}}$, as defined in Equation~\ref{loss:coo}. 2) Meanwhile, we employ the EDR loss to reshape the energy distribution for both ID image features and generated OOD image features, denoted as $\mathcal{L}_{\text{e}}(\mathbf{z_I})$ and $\mathcal{L}_{\text{e}}(\mathbf{z_I^c})$, respectively.
Therefore, the final optimization objective is expressed as:
\begin{equation}
    \mathcal{L}_{\text{CRoFT}} = \widehat{\mathcal{E}}_{\mathcal{S}}(\boldsymbol{\uptheta}) + \lambda_1 \mathcal{L}_{\text{c}} + \lambda_2 \left(\mathcal{L}_{\text{e}}(\mathbf{z_I}) + \mathcal{L}_{\text{e}}(\mathbf{z_I^c})\right)
\end{equation}
where $\lambda_1$ and $\lambda_2$ are hyperparameters that can be chosen based on
the validation procedure. For details about the complete algorithm, please refer to Algorithm~\ref{algoritm} in Appendix.


\begin{table*}[!t]
\centering
\vskip -0.1in
\caption{\textbf{Setup-I:} Comparison with competitive fine-tuning methods based on CLIP ViT-B/16. We report the average percentage results across 3 runs, with standard errors presented in parentheses. 
We can observe that CRoFT surpasses the state-of-the-art methods in distinguishing between closed-set OOD and open-set OOD samples, achieving a significant reduction of up to \textbf{25.3\%} on FPR95 when compared to CLIP. 
Meanwhile, CRoFT demonstrates comparable or even better generalization results on the closed-set OOD test sets.
}
\vskip 0.06in
\resizebox*{1.0\linewidth}{!}{
\begin{tabular}{c|cccc|cccc|cccc} 
\toprule
\multirow{2}{*}{\diagbox{Method}{DATA}} & \multicolumn{4}{c|}{1 -shot}                                                     & \multicolumn{4}{c|}{16-shot}                                                           & \multicolumn{4}{c}{32-shot}                                                            \\ 
\cline{2-13}
                                        & ID ACC        & OOD ACC             & AUROC               & FPR95               & ID ACC              & OOD ACC             & AUROC               & FPR95               & ID ACC              & OOD ACC             & AUROC               & FPR95                \\ 
\hline
CLIP                                    & 78.2          & 58.4                & 75.6                & 77.3                & 78.2                & 58.4                & 75.6                & 77.3                & 78.2                & 58.4                & 75.6                & 77.3                 \\
COOP                                    & 79.3          & 60.5                & 76.7                & 78.1                & 82.3                & 61.3                & 77.8                & 73.4                & 83.0                & 61.5                & 79.1                & 70.6                 \\
COCOOP                                  & \textbf{80.0} & 61.7                & 77.3                & 77.5                & 81.6                & 62.8                & 78.1                & 73.6                & 81.7                & 61.8                & 72.3                & 79.4                 \\
CLIP-Adapter                            & 78.2          & 58.2                & 75.9                & 79.3                & 78.7                & 59.0                & 76.0                & 79.7                & 78.6                & 59.1                & 76.1                & 79.5                 \\
Tip-Adapter-F                           & 79.5          & \textbf{61.9}                & 76.3                & 79.6                & 82.3                & 62.5                & 71.7                & 82.9                & \textbf{83.3}                & 62.8                & 68.1                & 85.9                 \\
DPLCLIP                                 & 79.1          & 60.4                & 77.9                & 77.3                & 82.1                & 61.7                & 78.6                & 72.4                & 83.0                & 61.7                & 77.8                & 72.6                 \\
Bayes-CAL                               & 79.0          & 60.5                & 75.5                & 77.2                & 82.1                & 61.3                & 78.3                & 71.3                & 82.9                & 61.5                & 78.3                & 70.9                 \\
\hline
\rowcolor[rgb]{0.941,0.941,0.941}  {\textbf{CRoFT~(Ours)}}                           & 79.6~\small(0.4)    & \textbf{61.9}~\small(0.3) & \textbf{80.5}~\small(1.2) & \textbf{69.3}~\small(3.5) & \textbf{82.5}~\small(0.2) & \textbf{62.9}~\small(0.6) & \textbf{87.2}~\small(0.9) & \textbf{52.0}~\small(3.3) & {83.1}~\small(0.5) & \textbf{63.1}~\small(0.5) & \textbf{86.5}~\small(0.3) & \textbf{55.2}~\small(0.8)  \\
\bottomrule
\end{tabular}
}
\label{tab:setup1_results}
\end{table*}

\section{Experiments}
\label{sec:exp}
In this section, we compare our method with competitive CLIP-based lightweight fine-tuning methods, such as CoOp \cite{zhou2021learning}, CoCoOp \cite{zhou2022cocoop}, CLIP-Adapter \cite{gao2023clip}, Tip-Adapter-F \cite{zhang2021tip}, DPLCLIP \cite{APCLIP} and Bayes-CAL \cite{zhu2023bayesian}. Additionally, we perform extensive ablation studies and visualization analyses to validate our theoretical findings. 
To better evaluate models' open-set OOD detection capabilities in real-world scenarios, we introduce two data settings that encompass 600 and 200+ unseen classes, respectively.
Details are illustrated as follows:

\textbf{Data setups}
1) \textit{Setup-I: open-set discrimination on the large-scale ImageNet dataset.} 
In the literature, ImageNet and its variants are commonly used for investigating fine-tuning robustness. In line with existing research, we construct datasets of Setup-I using these datasets.
Specifically, we split ImageNet-1K \cite{imagenet} into open and closed sets w.r.t class labels. We randomly define 40\% classes of ImageNet as the closed-set for training, and the remaining 60\% as the open-set for testing. The samples from ImageNet-A \cite{hendrycks2021natural}, ImageNet-R \cite{hendrycks2021many}, ImageNet-Sketch \cite{wang2019learning}, and ImageNet-V2 \cite{recht2019imagenet} with the same class labels with the closed-set are as closed-set OOD data.

2) \textit{Setup-II: open-set discrimination on cross-dataset images.} Using cross-dataset examples as the open-set is another
established protocol \cite{shafaei2018less, kong2021opengan}, which is introduced to reduce dataset-level bias. This protocol assesses the generalization capabilities of open-set OOD detection methods to diverse open-testing examples. 
In our experiment, we leverage popular datasets like PACS \cite{li2017deeper} or VLCS \cite{li2017deeper} for domain generalization studies as the closed-set data. We evaluate the models' ability to differentiate between closed-set OOD and cross-dataset images by utilizing different styles of datasets like Caltech101 \cite{bansal2021transfer}, DTD \cite{sharan2014accuracy}, and Food101 \cite{bossard2014food} as open-set examples. 
All overlapping classes are removed from the three open-set datasets.
In the evaluation of OOD generalization performance, we utilize the leave-one-domain-out validation protocol \cite{gulrajani2020search, cha2021swad} that uses three domains as closed-set ID data and the remaining one as closed-set OOD data.


\subsection{Experiment results of Setup-I}


\textbf{Experiment details} 
In our experiments on Setup-I, we conduct experiments based on the CLIP ViT-B/16 model.
For the prompt learning methods, CoOp, CoCoOp, DPLCLIP, and Bayes-CAL, we use random initialization for context vectors and set the number of context tokens to 16. 
Regarding other hyperparameters, we set the class token position (CTP) as ``end'' and set the class-specific context (CSC) as ``False''. This configuration has yielded the best average performance according to CoOp's paper. 
We adhere to the recommended hyperparameter settings outlined in the original paper of CLIP-Adapter.
In the case of Tip-Adapter-F, we perform hyperparameter searches following its original paper.
Without otherwise specified, methods are trained using the SGD optimizer with a learning rate of 0.002 and batch size of 32 for fair comparisons.
In our CRoFT method, we search for $\lambda_1$ in the range of $[1,5,10,15,20]$ and $\lambda_2$ in the range of $[10, 20, 30, 40, 50]$. 
As lightweight fine-tuning methods facilitate fast convergence, we set the maximum training epoch to 30. We compare our methods with these competitive fine-tuning methods under 1-shot, 16-shot and 32-shot scenarios.
For experiments on each method, we repeat 3 times with different random splits to eliminate the effects of randomness. Finally, we report the average classification accuracy on closed-set test sets, as well as the average FPR95 and AUROC results for distinguishing between open-set OOD data and closed-set data by inferring energy score \cite{liu2020energy}. For more experiment details, please refer to \cref{app:exp_details}.

\begin{figure}
\centering
\centerline{\includegraphics[width=0.98\linewidth]{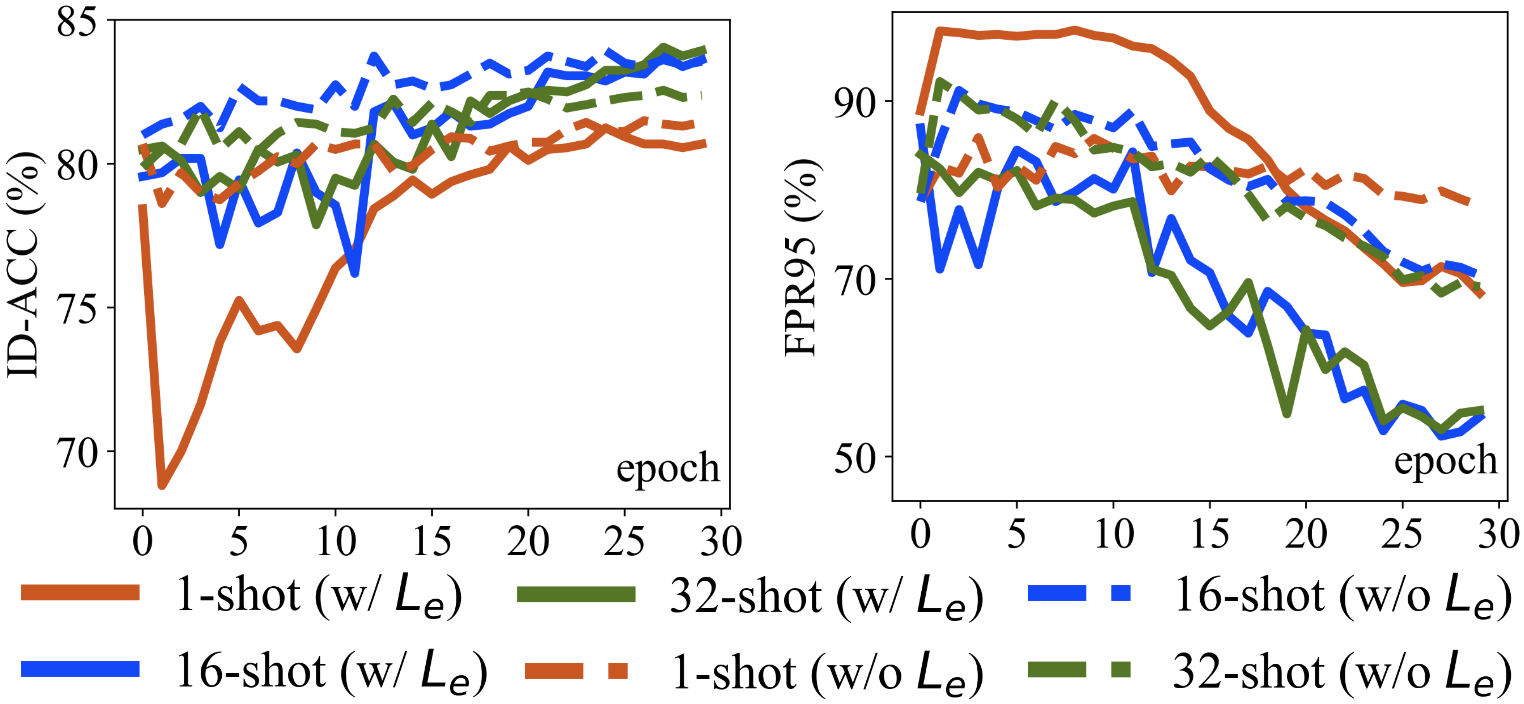}}
\vskip -0.1in
\caption{Ablations on the proposed EDR loss. With the proposed EDR loss $\mathcal{L}_{\text{e}}$, our method successfully fine-tunes CLIP's features in the direction of better-discriminating open-set and closed-set, without sacrificing the test accuracy. 
}
\vskip -0.1in
\label{fig:ablate_on_le}
\end{figure}

\textbf{Experiment results} 
We present the main results in Table~\ref{tab:setup1_results}, where CRoFT establishes the overall best performance in both OOD generalization and open-set OOD detection.
Notably, the proposed CRoFT method outperforms the zero-shot CLIP model by achieving an impressive improvement of up to \textbf{25.3\%} on FPR95. 
In contrast, its competitors struggle with the open-set OOD detection task, even resulting in higher FPR95 values when compared to the CLIP model.
Moreover, our method demonstrates comparable or even superior generalization results on the closed-set OOD test sets. Specifically, it achieves the best average test accuracy on closed-test OOD data compared to recent methods.

\begin{table}[!t]
\centering
\caption{Comparison with the vanilla technique of energy minimization (named EnergyMin). We report the classification accuracy on closed-set OOD data and the FPR95 value in open-set OOD detection (OOD ACC / FPR95).}
\vskip 0.06in
\resizebox*{1\linewidth}{!}{
\begin{tabular}{c|ccc} 
\toprule
{Method}               & {1-shot} & {16-shot} & {32-shot}  \\ 
\hline
CLIP~+~EnergyMin & \textbf{80.02} / 80.50             & 81.91 / 69.27              & 81.87 / 69.12               \\
CRoFT      & 79.60 / \textbf{69.31}             & \textbf{82.50 / 52.01}              & \textbf{83.10 / 55.19}               \\
\bottomrule
\end{tabular}
}
\label{tab:abalte_on_ErergyMin}
\vskip -0.15in
\end{table}

\begin{table}[!t]
\centering
\caption{Abaltion Study results on the proposed regularization loss $\mathcal{L}_{\text{c}}$ under 32-shot scenarios.}
\vskip 0.06in
\resizebox*{0.98\linewidth}{!}{
\begin{tabular}{c|cccccc} 
\toprule
    DATA       & ID    & INV2   & IN-S  & IN-A   & IN-R   & Avg OOD  \\ 
\hline
w~/~o~$\mathcal{L}_{\text{c}}$  & 82.21 & 75.83 & 42.98 & 69.57 & 60.57 & 62.24   \\
w~/~o~$\mathcal{L}_{e}$  & 81.99 & 76.33 & 43.55 & 70.18 & {61.06} & 62.78    \\
w~/~$\mathcal{L}_{\text{c}}$    & \textbf{83.10} & \textbf{76.64} & \textbf{44.23} & \textbf{70.30} & \textbf{61.15} & \textbf{63.07}    \\
\bottomrule
\end{tabular}
}
\label{tab:ablate_on_lc}
\vskip -0.1in
\end{table}

\textbf{Effect of $\mathcal{L}_{\text{e}}$} 
We ablate on the proposed EDR loss $\mathcal{L}_\text{e}$ and plot the corresponding ID test accuracy and open-set OOD detection results along the fine-tuning process. 
As depicted in Figure~\ref{fig:ablate_on_le}, the FPR95 value gradually decreases with the increase of training epoch, while maintaining the ID test accuracy. 
In Table~\ref{tab:abalte_on_ErergyMin}, we compare our method with the vanilla energy reduction technique (named EnergyMin). It is shown that this method achieved lower FPR95 values in the 16-shot and 32-shot scenarios compared to CLIP, but it still lagged behind our method by about 15\%. More importantly, our method demonstrated better OOD generalization results, thus validating our theoretical finding in Theorem~\ref{Th:2} that the EDR loss has a positive impact on OOD generalization.

\textbf{Effect of $\mathcal{L}_{\text{c}}$}~By removing the optimization process on the generated OOD image features while keeping other components unchanged, we assess the impact of $\mathcal{L}_{\text{c}}$ on OOD generalization. The results of this ablation study are in Table~\ref{tab:ablate_on_lc}.
With the employment of the $\mathcal{L}_{\text{c}}$, our method has obtained improved OOD test accuracy, especially in the worst-case scenario (i.e., ImageNet-Sketch), which highlights the efficacy of the proposed adversarial learning procedure in learning more robust adapters.


\begin{table*}[!t]
\centering
\vskip-0.1in
\caption{\textbf{Setup-II:} OOD generalization results measured by classification accuracy on closed-set OOD data (OOD ACC) and open-set OOD detection measured by AUROC and FPR95 over the mixture of closed-set OOD and open-set OOD test sets. 
For CLIP-Adapter and our method, the open-vs-closed discrimination results are obtained by inferring KNN distances ($k = 1$) or energy scores on the adapted image features. Due to space limitations, we only report the best results for each method. For the CLIP model, we also report its corresponding KNN-based results in CLIP (KNN).
Since other methods like CoOp, CoCoOp, and Tip-Adapter-F do not fine-tune the image features, their results when inferring KNN distances are the same as CLIP (KNN).
}
\vskip 0.06in
\resizebox*{1.0\linewidth}{!}{
\begin{tabular}{ccccccccccccccc}%
\toprule
\multicolumn{1}{c}{DATA} & \multirow{2}{*}{V-Net}    
& &\multirow{2}{*}{\makecell[c]{PACS\\OOD~ACC}}     &\multicolumn{3}{c}{PACS vs. Open-Set (AUROC~/~FPR95)} 
& & \multirow{2}{*}{\makecell[c]{VLCS\\OOD~ACC}}    &\multicolumn{3}{c}{VLCS vs. Open-Set (AUROC~/~FPR95)} 
& \multirow{2}{*}{\makecell[c]{\textbf{AVG}\\ \textbf{FPR}}}    
\\
\cmidrule{1-1}
\cmidrule{5-7}
\cmidrule{10-12}
\multicolumn{1}{c}{Algorithm} & & &  & DTD  & Food101 & Caltech101    & &  & DTD  & Food101 & Caltech101   \\
\hline 
\multirow{2}{*}{{CLIP}}   
       &\text{RN50}
       & & 90.8 \small(0.0) & 76.6~/~80.2 & 94.7~/~29.2 & 86.8~/~52.3 
       & & 75.1 \small(0.0) & 40.6~/~94.5 & 82.8~/~52.8 & 61.5~/~85.5 & 65.8
       \\
       &\text{ViT16}
       & & 96.1 \small(0.0) & 82.4~/~67.6 & 95.9~/~26.0 & 86.7~/~\textbf{52.2}
       & & 76.3 \small(0.0) & 55.3~/~88.8 & 85.8~/~48.3 & 53.3~/~86.3  & 61.5
       \\
\multirow{2}{*}{CLIP (KNN)}   &\text{RN50}
       & & 90.8 \small(0.0) & 85.3~/~58.5 & 99.1~/~4.3  & 79.2~/~65.3 
       & & 75.1 \small(0.0) & 82.4~/~67.4 & 95.2~/~27.0   & 74.1~/~74.4 & 49.9
       \\
       &\text{ViT16}
       & & 96.1 \small(0.0) & 87.7~/~58.9 & 98.8~/~5.8   & 82.5~/~70.0 
       & & 76.3 \small(0.0) & 82.4~/~71.6 & 95.1~/~32.6   & 78.7~/~74.1 & 52.2
       \\
\hline 
\multirow{2}{*}{CoOp} &\text{RN50} 
       & & 91.5 \small(0.6) & 90.1~/~43.6 & 97.0~/~15.5 & 83.1~/~56.8 
       & & 72.8 \small(5.4) & 40.4~/~96.9 & 54.5~/~87.0 & 46.3~/~93.7 & 65.6
         \\
       &\text{ViT16}
       & & 96.3 \small(0.7) & 85.7~/~62.7 & 96.2~/~22.6 & 80.0~/~70.9 
       & & 78.3 \small(1.7) & 47.3~/~89.2 & 76.3~/~62.9 & 46.8~/~90.0  & 66.4
       \\
\hline 
\multirow{2}{*}{CoCoOp} &\text{RN50} 
       & & 91.8 \small(0.6) & 90.7~/~42.8 & 98.3~/~9.0 & {88.4~/~48.5} 
       & & 76.3 \small(1.0) & 42.4~/~95.9 & 80.2~/~65.0 & 46.8~/~92.5 & 58.9
       \\
       &\text{ViT16}
       & & 96.8 \small(0.5)  & {92.1~/~44.9} & 97.8~/~15.0 & {87.8~/~57.5} 
       & & 78.9 \small(0.8) & 49.2~/~93.3 & 78.4~/~60.8 & 49.2~/~90.6  & 60.4
       \\
\hline 
\multirow{2}{*}{CLIP-Adapter} &\text{RN50}
       & & 90.9 \small(0.1) & 85.3~/~58.2 & 99.1~/~4.2   & 79.2~/~65.0
       & & 75.2 \small(0.1) & 82.4~/~67.7 & \textbf{95.3~/~26.5}   & 74.1~/~74.7 
       & 49.4 \\
       &\text{ViT16}
       & & 96.1 \small(0.0) & 87.7~/~58.6 & 98.8~/~\textbf{5.6}   & 82.6~/~69.6 
       & & 77.3 \small(0.7) & 82.6~/~71.2 & 95.1~/~32.4   & 78.8~/~74.2 
       & 51.9 \\
\hline 
\multirow{2}{*}{{Tip-Adapter-F}} &\text{RN50}
       & & 92.2 \small(0.4) & 83.4~/~81.8 & 97.2~/~15.1   & 84.7~/~72.9 
       & & 76.5 \small(0.2) & 59.3~/~95.4 & 91.5~/~38.6   & 64.1~/~84.0  & 65.7
       \\
       &\text{ViT16}
       & & 96.9 \small(0.4) & 87.3~/~69.3 & 97.8~/~11.2   & 84.8~/~74.9 
       & & 79.9 \small(0.2) & 66.7~/~94.4 & 92.7~/~39.6   & 63.0~/~95.1  & 64.1
       \\
\hline 
\multirow{2}{*}{DPLCLIP}      & RN50  & & 89.6 \small(1.2) & \textbf{95.4~/~25.0}                      & 99.0~/~5.8   & \textbf{97.4~/~14.1} 
             & & 76.1 \small(0.6) & 56.3 / 93.8                       & 79.9~/~67.6 & 51.5~/~96.0 & 50.4 \\
             & ViT16 & & 95.6 \small(0.2) & 93.2~/~39.0                      & 98.0~/~13.0 & \textbf{94.4~/~33.5} 
             & & 76.5 \small(1.3) & 52.9~/~98.7                       & 69.7~/~88.8 & 52.4~/~97.3 & 61.8 \\
\hline
\multirow{2}{*}{Bayes-CAL}    & RN50  & & 91.8 \small(0.3) & 89.7~/~47.8                      & 97.4~/~14.8 & 83.5~/~70.5
             & & 78.1 \small(1.5) & 50.9~/~94.3                       & 69.5~/~76.0 & 53.4~/~97.7 & 66.9 \\
             & ViT16 & & 96.6 \small(0.5) & 79.2~/~62.6                      & 87.0~/~38.6 & 71.8~/~60.3 
             & & 79.6 \small(0.9) & 44.2~/~93.6                       & 68.1~/~72.9 & 34.0~/~90.7 & 69.8 \\
\hline
\rowcolor[rgb]{0.941,0.941,0.941}  &\text{RN50}
       & & \textbf{92.5 \small(0.3)} & {94.7~/~26.8} & \textbf{99.8~/~0.8} & {88.7~/~42.0}
       & & \textbf{79.5 \small(0.7)} & \textbf{86.6~/~58.8} & 93.9~/~35.3 & \textbf{77.7~/~68.1} & \textbf{38.6}
       \\
       \rowcolor[rgb]{0.941,0.941,0.941}
       \multirow{-2}{*}{\textbf{CRoFT~(Ours)}}
       &\text{ViT16}
       & & \textbf{97.3 \small(0.1)} & \textbf{94.7~/~33.2} & \textbf{99.1}~/~6.1 & {88.9~/~52.2}
       & & \textbf{80.2 \small(1.0)} & \textbf{86.7~/~60.6} & \textbf{96.0~/~25.9}  & \textbf{83.7~/~62.2} & \textbf{40.0}
       \\
\bottomrule
\end{tabular}
}
\label{tab:setup2_results}
\vskip -0.1in
\end{table*}

\subsection{Experiment Results of Setup-II}

\textbf{Experiment details} 
In Setup-II, we keep the same hyperparameter setting as in Setup-I without further explanation.
For our CRoFT method, we set $\lambda_1$ as 15 and search for $\lambda_2$ in the range of $[1, 10, 100, 1000]$.
Following previous works \cite{zhou2022cocoop, zhu2023bayesian}, we perform evaluations under 16-shot scenarios based on CLIP RN50 and ViT-B/16.
All experiments are repeated 3 times with different random seeds. 
Finally, we report the average classification accuracy on closed-set OOD data, as well as the average FPR95 and AUROC for distinguishing between open-set OOD and closed-set OOD by inferring energy score \cite{liu2020energy} and KNN distances \cite{sun2022out}.

\begin{figure*}
\centering
\centerline{\includegraphics[width=1\linewidth]{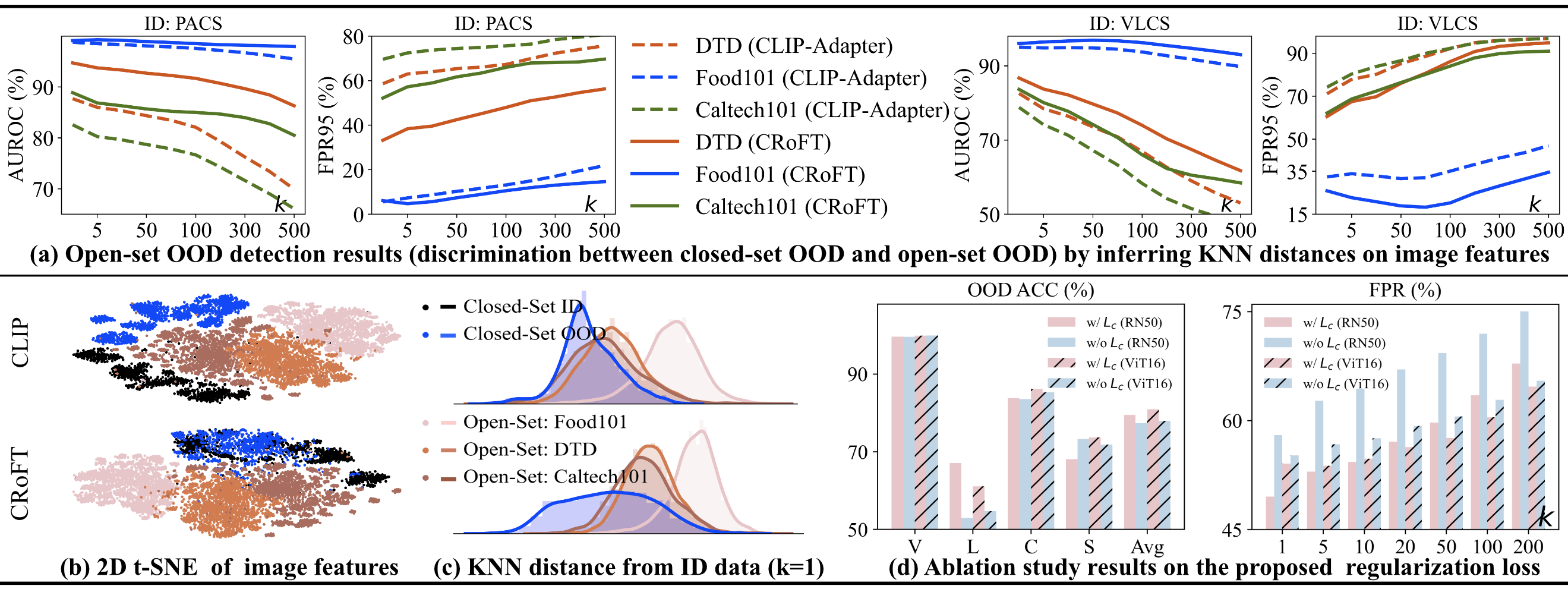}}
\vskip -0.1in
\caption{{Ablation Study results of Setup-II.} (a): Comparison with CLIP-Adapter in open-set OOD detection by referring KNN distances on the adapted image features. (b): t-SNE visualization of image features. (c): Average KNN distance between OOD features and ID features. (We use VLCS as the closed-set data in (b) and (c).)
(d): Experiment results on VLCS for CRoFT with $\mathcal{L}_{\text{c}}$ vs. without $\mathcal{L}_{\text{c}}$.}
\label{fig:ablate_on_setup2_fig}
\vskip -0.1in
\end{figure*}

\textbf{Experiment results} 
We present the main results in Table~\ref{tab:setup2_results}, where CRoFT establishes the overall best performance in both OOD generalization and open-set OOD detection.
The proposed CRoFT method outperforms the second-best FPR95 results by about \textbf{11\%}, showcasing its effectiveness in detecting cross-dataset unseen classes.
Consistent with the results of Setup-I, its competitors achieve even higher FPR95 results in comparison with the CLIP model.
It is worth noting that all methods yield better recognition results on the Food101 dataset but significantly worse results on the Caltech101 dataset. This difference can be attributed to the fact that Caltech101 is distributed closer to the in-distribution datasets. Despite the disparity degree between closed-set and open-set data, our method consistently improves performance across various pre-trained image encoders.
Moreover, our method demonstrates superior generalization results on the closed-set OOD test sets, 
boosting test accuracy on closed-test OOD data by an impressive \textbf{4.3\%} compared to the vanilla CLIP-Adapter method.

\textbf{Visualization of adapted features} 
To comprehensively illustrate that our CRoFT method can learn more distinct closed-set image features from open-set image features, we compare it with the vanilla adapter-tuning method (CLIP-Adapter) in open-set OOD detection by inferring KNN distances. As shown in Figure~\ref{fig:ablate_on_setup2_fig}~(a), our method consistently outperforms the CLIP-Adapter across various settings of $k$, showcasing the effectiveness and robustness of our method in learning discriminative features for open-vs-closed discrimination. 
Moreover, we visualize the distribution of image features in Figure~\ref{fig:ablate_on_setup2_fig}~(b)-(c). It is shown that image features from the zero-shot CLIP also contain domain-related information, which hinders the discrimination between closed-set OOD and open-set OOD data. In contrast, our method can learn open-vs-closed discriminated image features, achieving much smaller distances between the ID data and covariate-shifted OOD data.

\begin{table}[!t]
\centering
\vskip-0.1in
\caption{Ablation study results on $\mathcal{L}_{c}$ and $\mathcal{L}_{e}$. 
The best overall performances of our CRoFT validate the result of Theorem~\ref{Th:2}.}
\vskip 0.06in
\resizebox*{1.0\linewidth}{!}{
\begin{tabular}{cccccccc} 
\toprule
\multirow{2}{*}{DATA} & Method                      & \multicolumn{2}{c}{w~/~o $\mathcal{L}_{\text{c}}$}                         & \multicolumn{2}{c}{w~/~o $\mathcal{L}_{\text{e}}$}                         & \multicolumn{2}{c}{Ours}                               \\ 
\cline{2-8}
                      & V-Net                       & \multicolumn{1}{c}{RN50} & \multicolumn{1}{c}{ViT16} & \multicolumn{1}{c}{RN50} & \multicolumn{1}{c}{ViT16} & \multicolumn{1}{c}{RN50} & \multicolumn{1}{c}{ViT16}  \\ 
\hline
\multirow{2}{*}{PACS} & {OOD~ACC} & 92.29                    & 96.67                     & 92.02                    & 96.71                     & \textbf{92.45}                    & \textbf{97.26}                      \\
                      & AUROC                       & 88.28                    & 90.43                     & 91.49                    & 93.63                     & \textbf{94.41}                    & \textbf{94.25}                      \\
\hline
\multirow{2}{*}{VLCS} & OOD~ACC                        & 77.34                    & 77.95                     & 77.70                    & 77.56                     & \textbf{79.47}                    & \textbf{80.21}                      \\
                      & AUROC                       & 84.89                    & 86.50                     & 85.34                    & 88.31                     & \textbf{86.06}                    & \textbf{88.79}                      \\
\bottomrule
\end{tabular}
}
\label{tab:ablate_on_lc_2}
\vskip-0.2in
\end{table}

\textbf{Ablate on $\mathcal{L}_{c}$ and $\mathcal{L}_{e}$}
We conduct ablation studies on the proposed $\mathcal{L}_{c}$. As depicted in Figure~\ref{fig:ablate_on_setup2_fig}~(d), we observe that the OOD generalization results, especially for challenging examples (i.e., domain ``L''), are significantly enhanced by incorporating $\mathcal{L}_{\text{c}}$. Moreover, the open-set OOD detection results are also promoted, which is caused by the improved quality of the adapted image features.
Furthermore, we conduct additional ablation studies to validate the positive impact of the EDR loss on OOD generalization. We remove the EDR loss for all image features and evaluate the OOD generalization results as reported in Table~\ref{tab:ablate_on_lc_2}. It is shown that incorporating the EDR loss $\mathcal{L}_{\text{e}}$ leads to enhanced OOD generalization, highlighting its concurrent optimization effect on both OOD generalization and open-set OOD detection.

\section{Related Works}
\textbf{CLIP-based fine-tuning methods}
For training efficiency, there have been many lightweight CLIP-based fine-tuning methods to enhance generalization performance via prompt tuning \cite{singha2023ad, huang2022unsupervised, khattak2023maple, wang2023improving, wasim2023vita, zhu2023bayesian} or adapter tuning \cite{gondal2024domain, zhang2023unsupervised, zhu2024vision}. 
Prompt tuning methods aim to get better vision-language alignment via only fine-tuning the input prompts. For example, with only few-shot samples for learning, CoOp \cite{zhou2021learning} improved significantly in generalization ability over intensively-tuned manual prompts via prompt learning. Motivated by learning generalization prompts, CoCoOp \cite{zhou2022cocoop} is proposed to achieve generalization on unseen classes via conditional prompt learning. 
Adapter-tuning is another popular lightweight fine-tuning method, like CLIP-Adapter \cite{gao2023clip} and Tip-Adapter-F \cite{zhang2021tip}. Both of them inject a lightweight bottleneck architecture after the image encoder and perform residual-style feature blending with the original pre-trained embeddings.

\textbf{Open-set OOD detection}
There are multiple lines of work addressing open-set OOD detection, such as anomaly detection \cite{zong2018deep, liang2017enhancing}, outlier detection \cite{bevandic2021dense, saito2021openmatch}, and open-set OOD recognition \cite{kong2021opengan, geng2020recent, scheirer2014probability}. 
These methods can be categorized into two main groups: post hoc methods \cite{zhu2022boosting, liu2020energy, sun2021react, hendrycks2016baseline, liang2017enhancing, wang2022vim, sun2022out} and training-time regularization \cite{Narayanaswamy_2023_ICCV, bai2023feed, malinin2018predictive, du2022vos, du2022siren, ming2022poem}.
The former typically resort to post-hoc functions to recognize open-set without altering the DNN training process, like density estimation \cite{zhang2020hybrid}, uncertainty modeling \cite{gal2016dropout}, and input image reconstruction \cite{pidhorskyi2018generative, sun2020conditional}.
On the other hand, regularization-based methods aim to rectify the training process, compelling models to provide predictions with lower confidence. 


\section{Conclusion}
By connecting OOD generalization and OOD detection using Hessians, we have discovered that the proposed EDR loss not only approaches to minimizing energy scores on training data, but also leads to domain-consistent Hessians, thus enabling concurrent optimization for both OOD generalization and OOD detection.
Building upon this finding, we introduce a novel Hessian-based OOD generalization bound. From the perspective of bound minimization, we have developed a unified fine-tuning framework named CRoFT, aiming to enhance OOD generalization and open-set OOD detection simultaneously. Extensive experiments under different setups have demonstrated the superiority of our method. 


\section*{Acknowledgements}
This work is supported by National Natural Science Foundation of China under Grant (No.62106139) and State Key Laboratory of High Performance Computing, National University of Defense Technology (No.202401-11).

\section*{Impact Statement}
This paper presents work whose goal is to advance the field of robust fine-tuning of VL-PTMs. We aim to improve the model's capacity to address various distribution shifts in real-world applications. There are many potential societal consequences of our work, none of which we feel must be specifically highlighted here.




\newpage
\appendix
\onecolumn
\section{Proof of Theorem~\ref{Th:1}}
Before the proof, we first give an explanation of some important notations for clarity.
\begin{notation}
    We use domain $\mathcal{D}$ to represent a distribution on input space $\mathcal{X}$. Let $\widehat{\mathcal{D}}_\mathcal{T}$ and $\widehat{\mathcal{D}}_\mathcal{S}$ be the empirical distributions generated with $m$ i.i.d. samples from the source domain (training domain) and the target domain $\mathcal{D}_\mathcal{S}$ and training sets, respectively. Consider an bounded instance loss function $\ell$ such that $\mathcal{Y}\times\mathcal{Y} \rightarrow[0,c]$, and $\ell(\mathbf{y_1}, \mathbf{y_2})=0$ if and only if $\mathbf{y_1}=\mathbf{y_2}$ ($\mathbf{y_1} \in \mathcal{Y}, \mathbf{y_2} \in \mathcal{Y}$). We set $h$ as the true label function which generates the ground-truth label of inputs, i.e., $\mathbf{y}=h(\mathbf{x})$. Then we define functional error $\mathcal{E}_\mathcal{D}(f;h) := \mathbb{E}_\mathcal{D}(f(\mathbf{x};\mathbf{\boldsymbol{\uptheta}});h(\mathbf{x}))$, where $\boldsymbol{\uptheta}~(\boldsymbol{\uptheta}\in \Theta)$ is the parameter in predictor $f$. 
    Without introducing any ambiguity, we abbreviate the functional error as ${\mathcal{E}}_{\mathcal{D}}(\boldsymbol{\uptheta})$. Correspondingly,
    the empirical error calculated based on the empirical distribution $\widehat{\mathcal{D}}_\mathcal{S}$ ($\widehat{\mathcal{D}}_\mathcal{T}$) is denoted as $\widehat{\mathcal{E}}_\mathcal{S}(\boldsymbol{\uptheta})$ ($\widehat{\mathcal{E}}_\mathcal{T}(\boldsymbol{\uptheta})$); the empirical error on the covariate-shifted OOD data is $\widehat{\mathcal{E}}^c_\mathcal{S}(\boldsymbol{\uptheta})$ ($\widehat{\mathcal{E}}^c_\mathcal{T}(\boldsymbol{\uptheta})$).
\end{notation}

\begin{theorem}{\textbf{[Restatement of Theorem~\ref{Th:1}}]}
    Let the VC dimension of the parameter space $\Theta$ be $v$.
    Suppose that $\widehat{\mathcal{D}}_\mathcal{T}$ and $\widehat{\mathcal{D}}_\mathcal{S}$ are empirical distributions generated by drawing $m$ i.i.d. samples from the target and source domains, respectively.
    Denote the corresponding empirical errors by $\widehat{\mathcal{E}}_\mathcal{S}(\boldsymbol{\uptheta})$ and $\widehat{\mathcal{E}}_\mathcal{T}(\boldsymbol{\uptheta})$.
    Let ${\mathcal{E}}^c_\mathcal{S}(\boldsymbol{\uptheta})$ denote the error on covariate-shift data in the source domain.
    Let $\widehat{\mathcal{D}}_\mathcal{S}^c$ denote the empirical distribution of covariate-shift out-of-distribution data, and assume that it is close to the ID distribution, namely,
    $d_{\mathcal{H}\Delta\mathcal{H}}(\widehat{\mathcal{D}}_\mathcal{S}^c,\widehat{\mathcal{D}}_\mathcal{T}) \leq d_{\mathcal{H}\Delta\mathcal{H}}(\widehat{\mathcal{D}}_\mathcal{S},\widehat{\mathcal{D}}_\mathcal{T}) +\varepsilon_c$.
    Then, for $0 \leq \delta \leq 1$, with probability at least $1-\delta$, we have:
\begin{equation}
\begin{aligned}
&{\mathcal{E}}_\mathcal{T}({\boldsymbol{\uptheta}}) -\min_{\boldsymbol{\uptheta}^\prime} \mathcal{E}_\mathcal{T}({\boldsymbol{\uptheta}^\prime})
\leq \widehat{\mathcal{E}}_\mathcal{S}^c({\boldsymbol{\uptheta}})-\min_{\boldsymbol{\uptheta}^{\prime\prime}} {\mathcal{E}}_\mathcal{S}({\boldsymbol{\uptheta}^{\prime\prime}})
+\lambda   +\varepsilon_c
\\&+d_{\mathcal{H}\Delta\mathcal{H}}(\widehat{\mathcal{D}}_\mathcal{S},\widehat{\mathcal{D}}_\mathcal{T})
+O\left(\sqrt{\frac{v\log(m/v)+\log(1/\delta)}m}\right).
\end{aligned}
\label{eq:bound_step2}
\end{equation}
where $\lambda =\mathcal{E}_{\mathcal{S}}(\boldsymbol{\uptheta}^*) + \mathcal{E}_{\mathcal{T}}(\boldsymbol{\uptheta}^*)$, and $\boldsymbol{\uptheta}^*$ is the risk-minimizing optimum on the combined source and target data.

Using Taylor expansion, we express $\min_{\boldsymbol{\uptheta}^\prime} \mathcal{E}_\mathcal{T}({\boldsymbol{\uptheta}^\prime})$ in a Hessian-based form, namely,
$\min_{\boldsymbol{\uptheta}^\prime} 
\mathcal{E}_\mathcal{T}({\boldsymbol{\uptheta}^\prime}) 
\leq
{\mathcal{E}}_\mathcal{S}(\boldsymbol{\uptheta}^*)
+\frac12\left|\boldsymbol{\vartheta}^\top\mathbf{H}_\mathcal{S}(\boldsymbol{\theta}^*)\boldsymbol{\vartheta}\right| + \varepsilon  +
2\varepsilon_H$, where
$\boldsymbol{\vartheta}\in B_r(\mathbf 0)$.
Therefore, we obtain:
\begin{equation}
\begin{aligned}
{\mathcal{E}}_\mathcal{T}({\boldsymbol{\uptheta}})
&\leq \widehat{\mathcal{E}}_\mathcal{S}^c({\boldsymbol{\uptheta}})-\min_{\boldsymbol{\uptheta}^\prime}{\widehat{\mathcal{E}}}_\mathcal{S}({\boldsymbol{\uptheta}}^\prime)
+\frac12\left|{\boldsymbol{\vartheta}}^\top \mathbf{H}_\mathcal{S}({\boldsymbol{\uptheta}}^*){\boldsymbol{\vartheta}}\right|  +d_{\mathcal{H}\Delta\mathcal{H}}(\widehat{\mathcal{D}}_\mathcal{S},\widehat{\mathcal{D}}_\mathcal{T}) 
\\&+ \lambda_\mathcal{S}
+ \varepsilon  +
2\varepsilon_H +\varepsilon_c
+\max\{ \lambda_\mathcal{T}, 2 \lambda_\mathcal{S}- \lambda_\mathcal{T}\}
+O\left(\sqrt{\frac{v\log(m/v)+\log(1/\delta)}m}\right).
\end{aligned}
\end{equation}
where $\lambda =\mathcal{E}_{\mathcal{S}}(\boldsymbol{\uptheta}^*) + \mathcal{E}_{\mathcal{T}}(\boldsymbol{\uptheta}^*)$,
$
\varepsilon
:=
O(\varepsilon^\prime)
+
\frac{\rho}{6}r^3
$, $\varepsilon^\prime$ and $\varepsilon_c$ are small positive constants defined in Assumption~\ref{ass:xfinite2}, $\lambda_\mathcal{S} =\mathcal{E}_\mathcal{S}(\boldsymbol{\uptheta}^*)$, and $\lambda_\mathcal{T} =\mathcal{E}_\mathcal{T}(\boldsymbol{\uptheta}^*)$.
\end{theorem}

\textbf{Proof:}
The proof of Theorem~\ref{Th:1} consists of three parts.
First, we prove that the following inequality holds with high probability:
\begin{equation}
    \mathcal{E}_{\mathcal{T}}(\boldsymbol{\uptheta})
    \leq\widehat{\mathcal{E}}_{S}(\boldsymbol{\uptheta})+\frac12d_{\mathcal{H}\Delta\mathcal{H}}+O\left(\sqrt{\frac{v\log(m/v)+\log(1/\delta)}m}\right) +\lambda.
    \label{eq:bound_step1}
\end{equation}
Second, we connect this inequality to the covariate-shift OOD setting and obtain Eq.~\eqref{eq:bound_step2}.
Finally, we substitute the Hessian-based inequality, i.e., Theorem~\ref{Th:2}, into Eq.~\eqref{eq:bound_step2}.

Specifically, the first part relies on the following simple inequality, whose detailed proof can be found in Lemma 1 of previous work~\cite{cha2021swad}:
\begin{equation}
    \left|\mathcal{E}_\mathcal{S}(f,h)-\mathcal{E}_\mathcal{T}(f,h)\right|\leq\frac12d_{\mathcal{H}\Delta\mathcal{H}}(\mathcal{D}_\mathcal{S},\mathcal{D}_\mathcal{T})\quad 
    d_{\mathcal{H}\Delta\mathcal{H}}(\mathcal{D}_\mathcal{S},\mathcal{D}_\mathcal{T})=2{\sup}_A|{\Pr}_{\mathcal{D}_\mathcal{S}}(A)-{\Pr}_{\mathcal{D}_\mathcal{T}}(A)|  .
\end{equation}
By the triangle inequality, the following inequality holds with probability at least $1-\delta$:
\begin{equation}
\begin{aligned}
\mathcal{E}_\mathcal{T}(f,h^*)\quad
&\leq\quad\mathcal{E}_\mathcal{T}(h^*,h)+\mathcal{E}_\mathcal{T}(f,h^*)\leq\mathcal{E}_\mathcal{T}(h^*,h)+\mathcal{E}_\mathcal{S}(f,h^*)+|\mathcal{E}_\mathcal{T}(f,h^*)-\mathcal{E}_\mathcal{S}(f,h^*)|  \\
&\leq\quad\mathcal{E}_\mathcal{T}(h^*,h)+\mathcal{E}_S(f,h^*)+\frac12d_{\mathcal{H}\Delta\mathcal{H}}(\mathcal{D}_\mathcal{S},\mathcal{D}_\mathcal{T}) \\
&\leq\quad\mathcal{E}_\mathcal{T}(h^*,h)+\mathcal{E}_\mathcal{S}(f,h^*)+\mathcal{E}_S(h^*, h)+\frac12d_{\mathcal{H}\Delta\mathcal{H}}(\mathcal{D}_\mathcal{S},\mathcal{D}_T) \\
&=\quad\mathcal{E}_\mathcal{S}(f,h^*)+\frac12d_{\mathcal{H}\Delta\mathcal{H}}(\mathcal{D}_\mathcal{S},\mathcal{D}_\mathcal{T})+\lambda  \\
&\leq\quad\widehat{\mathcal{E}}_\mathcal{S}(f,h^*)+\frac12d_{\mathcal{H}\Delta\mathcal{H}}(\widehat{\mathcal{D}}_\mathcal{S},\widehat{\mathcal{D}}_\mathcal{T})+\sqrt{\frac{v(\log(m/v)+1)+\log(1/\delta)}{2m}}+\lambda .
\end{aligned}
\end{equation}
where $\lambda =\mathcal{E}_{\mathcal{S}}(h^*, h) + \mathcal{E}_{\mathcal{T}}(h^*, h)=\mathcal{E}_{\mathcal{S}}(\boldsymbol{\uptheta}^*) + \mathcal{E}_{\mathcal{T}}(\boldsymbol{\uptheta}^*)$, and $h^*=f(\cdot, {\boldsymbol{\uptheta}}^*)$ is the optimum on the combined source and target dataset.
The last step of the above inequality applies Lemma 2 in~\cite{blitzer2007learning}.
Considering the proposed adversarial-learning-based fine-tuning process, assume that $d_{\mathcal{H}\Delta\mathcal{H}}(\widehat{\mathcal{D}}_\mathcal{S}^c,\widehat{\mathcal{D}}_\mathcal{T}) \leq d_{\mathcal{H}\Delta\mathcal{H}}(\widehat{\mathcal{D}}_\mathcal{S},\widehat{\mathcal{D}}_\mathcal{T}) +\varepsilon_c$.
Then it is straightforward to show that:
\begin{equation}
\begin{aligned}
\mathcal{E}_\mathcal{T}({\boldsymbol{\uptheta}})
&\leq\widehat{\mathcal{E}}_\mathcal{S}^c({\boldsymbol{\uptheta}})+\frac12d_{\mathcal{H}\Delta\mathcal{H}}(\widehat{\mathcal{D}}_\mathcal{S},\widehat{\mathcal{D}}_\mathcal{T})+O\left(\sqrt{\frac{v\log(m/v)+\log(1/\delta)}{2m}}\right)+\lambda +\varepsilon_c.
\end{aligned}
\label{eq:ineq1}
\end{equation}

For the second part, let $\bar{{\boldsymbol{\uptheta}}}\in\operatorname{argmin}_{{\boldsymbol{\uptheta}}\in\Theta} {\mathcal{E}}_\mathcal{T}({\boldsymbol{\uptheta}})$.
Again using Lemma 2 in~\cite{blitzer2007learning}, the following inequality holds with probability at most $\delta$:
\begin{equation}
    |\widehat{\mathcal{E}}_\mathcal{S}(\bar{{\boldsymbol{\uptheta}}})-\mathcal{E}_\mathcal{S}(\bar{{\boldsymbol{\uptheta}}})|> O\left(\sqrt{\frac{v\log(m/v)+\log(1/\delta)}{2m}}\right).
\end{equation}
Moreover, we have the following inequality:
\begin{equation}
    |\mathcal{E}_\mathcal{T}({\boldsymbol{\uptheta}})-\mathcal{E}_{\mathcal{S}}({\boldsymbol{\uptheta}})|\leq \frac12d_{\mathcal{H}\Delta\mathcal{H}}(\widehat{\mathcal{D}}_\mathcal{S},\widehat{\mathcal{D}}_\mathcal{T}) + O\left(\sqrt{\frac{v\log(m/v)+\log(1/\delta)}{2m}}\right).
\end{equation}
Thus, with probability greater than $1-\delta$, the following inequality holds:
\begin{equation}
\begin{aligned}
\min_{{\boldsymbol{\uptheta}}^{\prime}}\widehat{\mathcal{E}}_{\mathcal{S}}({\boldsymbol{\uptheta}}^{\prime})
& \leq\widehat{\mathcal{E}}_{\mathcal{S}}(\bar{{\boldsymbol{\uptheta}}})\leq\mathcal{E}_{\mathcal{S}}^c(\bar{\boldsymbol{\uptheta}})+O\left(\sqrt{\frac{v\log(m/v)+\log(1/\delta)}{m}}\right)  \\
&\leq{\mathcal{E}}_\mathcal{T}(\bar{{\boldsymbol{\uptheta}}})+\frac12d_{\mathcal{H}\Delta\mathcal{H}}(\widehat{\mathcal{D}}_\mathcal{S},\widehat{\mathcal{D}}_T)+O\left(\sqrt{\frac{v\log(m/v)+\log(1/\delta)}{m}}\right)  \\
&=\min_{{\boldsymbol{\uptheta}}^{\prime}}\mathcal{E}_\mathcal{T}({\boldsymbol{\uptheta}}^{\prime})+\frac12d_{\mathcal{H}\Delta\mathcal{H}}(\widehat{\mathcal{D}}_\mathcal{S},\widehat{\mathcal{D}}_\mathcal{T})+O\left(\sqrt{\frac{v\log(m/v)+\log(1/\delta)}{m}}\right) .
\end{aligned}
\label{eq:ineq2}
\end{equation}

Substituting Eq.~\eqref{eq:ineq2} into Eq.~\eqref{eq:ineq1}, the OOD generalization bound can be written as:
\begin{equation}
\begin{aligned}
{\mathcal{E}}_{\mathcal{T}}({\boldsymbol{\uptheta}}) -\min_{{\boldsymbol{\uptheta}}^{\prime}} {\mathcal{E}}_\mathcal{T}({\boldsymbol{\uptheta}}^{\prime})
&\leq \widehat{\mathcal{E}}_\mathcal{S}^c({\boldsymbol{\uptheta}})-\min_{{\boldsymbol{\uptheta}}^{\prime}} \widehat{\mathcal{E}}_\mathcal{S}({{\boldsymbol{\uptheta}}^{\prime}})
+\lambda
\\&+d_{\mathcal{H}\Delta\mathcal{H}}(\widehat{\mathcal{D}}_\mathcal{S},\widehat{\mathcal{D}}_\mathcal{T})
+O\left(\sqrt{\frac{v\log(m/v)+\log(1/\delta)}m}\right).
\end{aligned}
\label{eq:bound_step_2}
\end{equation}

Finally, by the curvature-control condition of EDR, for any
$\boldsymbol{\vartheta}\in B_r(\mathbf 0)$, we have
\begin{equation}
\frac12
\left|
\boldsymbol{\vartheta}^\top
\mathbf C_{\mathcal D}(\boldsymbol{\uptheta}^\ast)
\boldsymbol{\vartheta}
\right|
\le
\varepsilon_H,
\qquad
\mathcal D\in\{\mathcal S,\mathcal T\}.
\end{equation}
Using Taylor expansion, we express $\min_{\boldsymbol{\uptheta}^\prime} \mathcal{E}_\mathcal{T}({\boldsymbol{\uptheta}^\prime})$ in a Hessian-based form:
$\min_{\boldsymbol{\uptheta}^\prime} 
\mathcal{E}_\mathcal{T}({\boldsymbol{\uptheta}^\prime}) 
\leq
\max\{{\mathcal{E}}_\mathcal{T}(\boldsymbol{\uptheta}^*), 2{\mathcal{E}}_\mathcal{S}(\boldsymbol{\uptheta}^*)-{\mathcal{E}}_\mathcal{T}(\boldsymbol{\uptheta}^*)\} 
+\frac12\left|\boldsymbol{\vartheta}^\top (\mathbf{H}_\mathcal{S}(\boldsymbol{\uptheta}^*))\boldsymbol{\vartheta}\right| + 
\varepsilon
+
2\varepsilon_H$.
This inequality follows directly from Theorem~\ref{Th:2}, whose details are provided in Appendix~\ref{appendix:proof_pf_Th2}.

Therefore, the OOD generalization bound in Eq.~\eqref{eq:bound_step_2} can be further written as:
\begin{equation}
\begin{aligned}
{\mathcal{E}}_\mathcal{T}(\boldsymbol{\uptheta})
&\leq \widehat{\mathcal{E}}_\mathcal{S}^c(\boldsymbol{\uptheta})-\min_{{\boldsymbol{\uptheta}}^{\prime}} \widehat{\mathcal{E}}_\mathcal{S}({{\boldsymbol{\uptheta}}^{\prime}})
+\frac12\left|\boldsymbol{\vartheta}^\top (\mathbf{H}_\mathcal{S}(\boldsymbol{\uptheta}^*))\boldsymbol{\vartheta}\right| + \varepsilon  +
2\varepsilon_H +\varepsilon_c
+d_{\mathcal{H}\Delta\mathcal{H}}(\widehat{\mathcal{D}}_\mathcal{S},\widehat{\mathcal{D}}_\mathcal{T}) 
\\&+ \lambda +  
\max\{ \lambda_\mathcal{T}, 2 \lambda_\mathcal{S}- \lambda_\mathcal{T}\}
+O\left(\sqrt{\frac{v\log(m/v)+\log(1/\delta)}m}\right).
\end{aligned}
\end{equation}

Therefore, the proof is completed.

\section{Proof of Lemma~\ref{lemma:note_for_bound}}

\textbf{Proof:}
To solve for the optimum of ${\min E_{\boldsymbol{\uptheta}}(\mathbf{x})}$, i.e.,
${\nabla_{\boldsymbol{\uptheta}} E_{\boldsymbol{\uptheta}}(\mathbf{x})} \rightarrow \boldsymbol{0}$,
we propose to minimize the squared norm of Eq.~\eqref{eq:gradient}, namely the norm of the gradient vector defined in Eq.~\eqref{eq:gradient}. The objective can be written as:
\begin{equation}
\mathcal{L}_{\text{e}}=\frac1N\sum_{i=1}^N\left[\nabla_{\boldsymbol{\uptheta_l}}\left(\log\sum_{j=1}^K\exp\left<\mathbf{{\mathbf{z_I}(\mathbf{x^{(i)}};\boldsymbol{\uptheta})}, z_T^{(j)}}(\boldsymbol{\uptheta})\right>\right)\right]^2.
\end{equation}
Here, ${\boldsymbol{\uptheta}} = \{\boldsymbol{\uptheta}_0, \boldsymbol{\uptheta}_l\}$, where ${\boldsymbol{\uptheta}}_0$ denotes the frozen parameters in the pretrained encoder, and ${\boldsymbol{\uptheta}}_l$ denotes the parameters of the linear projection layer to be optimized.

We now expand $\mathcal{L}_{\text{e}}$ as follows:
\begin{equation}
\begin{aligned}
\mathcal{L}_{\text{e}}&=\frac1N\sum_{i=1}^N\left[\nabla_{\boldsymbol{\uptheta_l}}\left(\log\sum_{j=1}^K\exp\left<\mathbf{{\mathbf{z_I}(\mathbf{x^{(i)}};\boldsymbol{\uptheta})}, z_T^{(j)}}(\boldsymbol{\uptheta})\right>\right)\right]^2 \\
& = \frac1N\sum_{i=1}^N\left[\frac{\nabla_{\boldsymbol{\uptheta}_l}\sum_{j=1}^K\exp\left<\mathbf{{\mathbf{z_I}(\mathbf{x^{(i)}};\boldsymbol{\uptheta})}, z_T^{(j)}}(\boldsymbol{\uptheta})\right>}{\sum_{j=1}^K\exp\left<\mathbf{{\mathbf{z_I}(\mathbf{x^{(i)}};\boldsymbol{\uptheta})}, z_T^{(j)}}(\boldsymbol{\uptheta})\right>}\right]^2. \\
\end{aligned}
\end{equation}

We denote the above expression as the squared norm of a vector $\mathbf{a}$, namely $\mathcal{L}_{\text{e}}=|\mathbf{a}|^2$, where
$$\mathbf{a} = -\frac1N\sum_{i=1}^N 
        \left[\frac{\nabla_{\boldsymbol{\uptheta}_l}\sum_{j=1}^K\exp\left<\mathbf{{\mathbf{z_I}(\mathbf{x^{(i)}};\boldsymbol{\uptheta})}, z_T^{(j)}}(\boldsymbol{\uptheta})\right>}{\sum_{j=1}^K\exp\left<\mathbf{{\mathbf{z_I}(\mathbf{x^{(i)}};\boldsymbol{\uptheta})}, z_T^{(j)}}(\boldsymbol{\uptheta})\right>}\right].$$

The empirical classification loss $\widehat{\mathcal{E}}_\mathcal{D}(\mathbf{z};\boldsymbol{\uptheta})$ is computed as:
\begin{equation}
\begin{aligned}
    \widehat{\mathcal{E}}_\mathcal{D}(\mathbf{z};\boldsymbol{\uptheta}) &= -\frac1N\sum_{i=1}^N \log\frac{\exp S^{(i)}}{\sum_{j=1}^K \exp\left<\mathbf{{\mathbf{z_I}(\mathbf{x^{(i)}};\boldsymbol{\uptheta})}, z_T^{(j)}}(\boldsymbol{\uptheta})\right>} \\
    &=\frac1N\sum_{i=1}^N \left[\log\sum_{j=1}^K\exp{\left<\mathbf{{\mathbf{z_I}(\mathbf{x^{(i)}};\boldsymbol{\uptheta})}, z_T^{(j)}}(\boldsymbol{\uptheta})\right>} -S^{(i)}\right].\\
\end{aligned}
\end{equation}

Accordingly, the gradient vector of the empirical risk $\widehat{\mathcal{E}}_\mathcal{D}(\mathbf{z};\boldsymbol{\uptheta})$ with respect to the parameter $\boldsymbol{\uptheta}_l$ can be written as:
\begin{equation}
    \begin{aligned}
        \widehat{\mathbf{G}}_\mathcal{D}(\boldsymbol{\uptheta}_l)
    &=\nabla_{\boldsymbol{\uptheta}_l}\widehat{\mathcal{E}}_\mathcal{D}(\boldsymbol{\uptheta})
        =\frac1N\sum_{i=1}^N \left[\frac{\nabla_{\boldsymbol{\uptheta}_l}\sum_{j=1}^K\exp\left<\mathbf{{\mathbf{z_I}(\mathbf{x^{(i)}};\boldsymbol{\uptheta})}, z_T^{(j)}}(\boldsymbol{\uptheta})\right>}{\sum_{j=1}^K\exp\left<\mathbf{{\mathbf{z_I}(\mathbf{x^{(i)}};\boldsymbol{\uptheta})}, z_T^{(j)}}(\boldsymbol{\uptheta})\right>} -\nabla_{\mathbf{\uptheta}_l}S^{(i)}\right] \\
        &= -\mathbf{a}-\frac1N\sum_{i=1}^N \nabla_{\mathbf{\uptheta}_l}S^{(i)}.
    \end{aligned}
\end{equation}

The Hessian matrix of the empirical risk $\mathcal{E}_\mathcal{D}(\boldsymbol{\uptheta})$ with respect to the parameter $\boldsymbol{\uptheta}_l$ is given by:
\begin{equation}
    \begin{aligned}
        \widehat{\mathbf{H}}_\mathcal{D}(\boldsymbol{\uptheta}_l)&=\nabla^2_{\boldsymbol{\uptheta}_l}\widehat{\mathcal{E}}_\mathcal{D}(\boldsymbol{\uptheta})=-\nabla_{\boldsymbol{\uptheta}_l}\mathbf{a} -\frac1N\sum_{i=1}^N \nabla^2_{\mathbf{\uptheta}_l}S^{(i)}.
    \end{aligned}
\end{equation}

Let
\begin{equation}
\mathbf{C}_\mathcal{D}(\boldsymbol{\uptheta}_l)
=
-\nabla_{\boldsymbol{\uptheta}_l}
\mathbf{a}(\boldsymbol{\uptheta}_l),
\qquad
\mathbf{M}_\mathcal{D}(\boldsymbol{\uptheta}_l)
=
-\frac1N
\sum_{i=1}^N
\nabla_{\boldsymbol{\uptheta}_l}^2
S^{(i)}(\boldsymbol{\uptheta}).
\end{equation}

Then we obtain the following exact decomposition:
\begin{equation}
\widehat{\mathbf{H}}_\mathcal{D}(\boldsymbol{\uptheta}_l)
=
\mathbf{C}_\mathcal{D}(\boldsymbol{\uptheta}_l)
+
\mathbf{M}_\mathcal{D}(\boldsymbol{\uptheta}_l).
\end{equation}

Next, we further analyze the EDR optimization objective corresponding to Eq.~\eqref{eq:loss_edr}. Let
\begin{equation}
g_i(\boldsymbol{\uptheta}_l)
=
\nabla_{\boldsymbol{\uptheta}_l}
A^{(i)}(\boldsymbol{\uptheta}),
\end{equation}
where
\begin{equation}
A^{(i)}(\boldsymbol{\uptheta})
=
\log
\sum_{j=1}^K
\exp
\left<
\mathbf{z_I}(\mathbf{{x}^{(i)}};\boldsymbol{\uptheta}),
\mathbf{{z_T}^{(j)}}(\boldsymbol{\uptheta})
\right>.
\end{equation}

Then the EDR loss can be written as:
\begin{equation}
\mathcal{L}_{\text{e}}
=
\frac1N
\sum_{i=1}^N
\|g_i(\boldsymbol{\uptheta}_l)\|_2^2.
\end{equation}

Taking its gradient gives:
\begin{equation}
\nabla_{\boldsymbol{\uptheta}_l}
\mathcal{L}_{\text{e}}
=
\frac2N
\sum_{i=1}^N
\mathbf{J}_i(\boldsymbol{\uptheta}_l)^\top
g_i(\boldsymbol{\uptheta}_l),
\end{equation}
where
\begin{equation}
\mathbf{J}_i(\boldsymbol{\uptheta}_l)
=
\nabla_{\boldsymbol{\uptheta}_l}^2
A^{(i)}(\boldsymbol{\uptheta}).
\end{equation}
\begin{equation}
\mathbf C_{\mathcal D}(\boldsymbol{\uptheta}_l)
:=
-\nabla_{\boldsymbol{\uptheta}_l}
\mathbf{a}(\boldsymbol{\uptheta}_l)=
\frac1N
\sum_{i=1}^N
\nabla_{\boldsymbol{\uptheta}_l}^2
A^{(i)}(\boldsymbol{\uptheta}).
\end{equation}

Therefore, a local optimum of Eq.~\eqref{eq:loss_edr} satisfies:
\begin{equation}
\sum_{i=1}^N
\mathbf{J}_i(\boldsymbol{\uptheta}_l)^\top
g_i(\boldsymbol{\uptheta}_l)
=\mathbf{0}.
\end{equation}
That is, the EDR loss compresses
$\nabla_{\boldsymbol{\uptheta}_l}^2
A^{(i)}(\boldsymbol{\uptheta})$
along the gradient direction
$\nabla_{\boldsymbol{\uptheta}_l}
A^{(i)}(\boldsymbol{\uptheta})$,
and controls the curvature variation of
$\mathbf C_{\mathcal D}(\boldsymbol{\uptheta}_l)$
in the gradient direction.

Furthermore, assume that the spectral norm of the curvature term $\mathbf C_{\mathcal D}(\boldsymbol{\uptheta}^\ast)$ satisfies:
\begin{equation}
\begin{aligned}
\left\|
\mathbf C_{\mathcal D}(\boldsymbol{\uptheta}^\ast)
\right\|_2
&=
\left\|
\frac1N
\sum_{i=1}^N
\nabla_{\boldsymbol{\uptheta}}^2
A^{(i)}(\boldsymbol{\uptheta}^\ast)
\right\|_2
\le
\frac1N
\sum_{i=1}^N
\left\|
\nabla_{\boldsymbol{\uptheta}}^2
A^{(i)}(\boldsymbol{\uptheta}^\ast)
\right\|_2 \le
L_A.
\end{aligned}
\end{equation}

For any
\(
\boldsymbol{\vartheta}\in B_r(\mathbf 0)
\),
by the quadratic-form inequality, we have:
\begin{equation}
\begin{aligned}
\left|
\boldsymbol{\vartheta}^\top
\mathbf C_{\mathcal D}(\boldsymbol{\uptheta}^\ast)
\boldsymbol{\vartheta}
\right|
&\le
\left\|
\mathbf C_{\mathcal D}(\boldsymbol{\uptheta}^\ast)
\right\|_2
\cdot
\|
\boldsymbol{\vartheta}
\|_2^2\le
L_A
\|
\boldsymbol{\vartheta}
\|_2^2\le
L_A r^2.
\end{aligned}
\end{equation}

That is,
\begin{equation}
\frac12
\left|
\boldsymbol{\vartheta}^\top
\mathbf C_{\mathcal D}(\boldsymbol{\uptheta}^\ast)
\boldsymbol{\vartheta}
\right|
\le
\frac12
L_A r^2
=\varepsilon_H.
\end{equation}

In the adapter-tuning scenario based on CLIP pretrained features, the inner product between the image feature and the text feature corresponding to the ground-truth class name is:
\begin{equation}
S^{(i)}
=
\left<
\boldsymbol{\uptheta}_\mathbf{I}
\mathbf{z_{I0}^{(i)}},
\boldsymbol{\uptheta}_\mathbf{T}
\mathbf{z_{T0}^{(i)}}
\right>.
\end{equation}

Here,
$\boldsymbol{\uptheta}_\mathbf{I}\in\mathbb{R}^{d\times d}$
and
$\boldsymbol{\uptheta}_\mathbf{T}\in\mathbb{R}^{d\times d}$
denote the image-adapter and text-adapter parameters, respectively.

Since $S^{(i)}$ is bilinear with respect to
$\boldsymbol{\uptheta}_\mathbf{I}$
and
$\boldsymbol{\uptheta}_\mathbf{T}$,
the diagonal blocks of its second derivative are zero, while the off-diagonal blocks are given by the cross term
$\mathbf{z_{I0}^{(i)}}\mathbf{z_{T0}^{(i)}}^\top$.
That is,
\begin{equation}
\mathbf{M}_\mathcal{D}(\boldsymbol{\uptheta}_l)
=
-\frac1N
\sum_{i=1}^N
\begin{bmatrix}
\mathbf{0}
&\mathbf{z_{I0}^{(i)}}\mathbf{z_{T0}^{(i)}}^\top
\\
\mathbf{z_{I0}^{(i)}}\mathbf{z_{T0}^{(i)}}^\top
&\mathbf{0}
\end{bmatrix}.
\end{equation}

In summary, let
$B_r(\mathbf{0})
=\left\{\boldsymbol{\vartheta}\in
\mathbb{R}^d : \| \boldsymbol{\vartheta} \|_2 \le r
\right\}$
denote the Euclidean closed ball centered at the origin $\mathbf{0}$ with radius $r$.
Under the curvature-control condition
\begin{equation}
\frac12
\left|
\boldsymbol{\vartheta}^{\top}
\mathbf C_{\mathcal D}(\boldsymbol{\uptheta}^{\ast})
\boldsymbol{\vartheta}
\right|
\le
\varepsilon_H,
\qquad
\forall \boldsymbol{\vartheta}\in B_r(\mathbf 0),
\end{equation}
the dominant term of the Hessian $\widehat{\mathbf{H}}_\mathcal{D}(\boldsymbol{\uptheta}_l)$ is determined by the domain-consistent second-order term
$\mathbf M_{\mathcal D}(\boldsymbol{\uptheta}_l)$.

Therefore, the proof is completed.

\section{Proof of Theorem~\ref{Th:2}}
\label{appendix:proof_pf_Th2}

\textbf{Proof:}
Let $\boldsymbol{\uptheta}^\ast$ be a local minimum of the empirical risk, i.e.,
$\nabla_{\boldsymbol{\uptheta}}
{\mathcal E}_{\mathcal D}(\boldsymbol{\uptheta}^\ast)
=\mathbf 0$,
$\mathcal D\in\{\mathcal S,\mathcal T\}$.
Take an arbitrary local perturbation
$\boldsymbol{\vartheta}\in B_r(\mathbf 0)$.
Since the Hessian of the population risk is $\rho$-Lipschitz continuous, applying the second-order Taylor expansion to the target-domain risk
${\mathcal E}_{\mathcal T}$
at
$\boldsymbol{\uptheta}^\ast$
gives:
\begin{equation}
{\mathcal E}_{\mathcal T}(\boldsymbol{\uptheta}^\ast+\boldsymbol{\vartheta})
=
{\mathcal E}_{\mathcal T}(\boldsymbol{\uptheta}^\ast)
+
\frac12
\boldsymbol{\vartheta}^\top
{\mathbf H}_{\mathcal T}(\boldsymbol{\uptheta}^\ast)
\boldsymbol{\vartheta}
+
R_{\mathcal T}(\boldsymbol{\vartheta}),
\end{equation}
where the remainder term satisfies:
\begin{equation}
|R_{\mathcal T}(\boldsymbol{\vartheta})|
\le
\frac{\rho}{6}
\|\boldsymbol{\vartheta}\|_2^3
\le
\frac{\rho}{6}r^3.
\end{equation}

Therefore,
\begin{equation}
\begin{aligned}
&
\left|
{\mathcal E}_{\mathcal T}(\boldsymbol{\uptheta}^\ast+\boldsymbol{\vartheta})
-
{\mathcal E}_{\mathcal S}(\boldsymbol{\uptheta}^\ast)
\right|
\le
\left|
{\mathcal E}_{\mathcal T}(\boldsymbol{\uptheta}^\ast)
-
{\mathcal E}_{\mathcal S}(\boldsymbol{\uptheta}^\ast)
\right|
+
\frac12
\left|
\boldsymbol{\vartheta}^\top
{\mathbf H}_{\mathcal T}(\boldsymbol{\uptheta}^\ast)
\boldsymbol{\vartheta}
\right|
+
\frac{\rho}{6}r^3.
\end{aligned}
\label{eq:proof_taylor}
\end{equation}

According to Lemma~\ref{lemma:note_for_bound}, the empirical-risk Hessian admits the following exact decomposition:
\begin{equation}
{\mathbf H}_{\mathcal D}(\boldsymbol{\uptheta}^\ast)
=
\mathbf C_{\mathcal D}(\boldsymbol{\uptheta}^\ast)
+
\mathbf M_{\mathcal D}(\boldsymbol{\uptheta}^\ast),
\qquad
\mathcal D\in\{\mathcal S,\mathcal T\}.
\end{equation}

Thus,
\begin{equation}
\begin{aligned}
{\mathbf H}_{\mathcal T}(\boldsymbol{\uptheta}^\ast)
=
{\mathbf H}_{\mathcal S}(\boldsymbol{\uptheta}^\ast)
&+
\Bigl(
\mathbf C_{\mathcal T}(\boldsymbol{\uptheta}^\ast)
-
\mathbf C_{\mathcal S}(\boldsymbol{\uptheta}^\ast)
\Bigr)
+
\Bigl(
\mathbf M_{\mathcal T}(\boldsymbol{\uptheta}^\ast)
-
\mathbf M_{\mathcal S}(\boldsymbol{\uptheta}^\ast)
\Bigr).
\end{aligned}
\end{equation}

It follows that
\begin{equation}
\begin{aligned}
&
\frac12
\left|
\boldsymbol{\vartheta}^\top
{\mathbf H}_{\mathcal T}(\boldsymbol{\uptheta}^\ast)
\boldsymbol{\vartheta}
\right|
\le
\frac12
\left|
\boldsymbol{\vartheta}^\top
{\mathbf H}_{\mathcal S}(\boldsymbol{\uptheta}^\ast)
\boldsymbol{\vartheta}
\right|
+
\frac12
\left|
\boldsymbol{\vartheta}^\top
\Bigl(
\mathbf C_{\mathcal T}(\boldsymbol{\uptheta}^\ast)
-
\mathbf C_{\mathcal S}(\boldsymbol{\uptheta}^\ast)
\Bigr)
\boldsymbol{\vartheta}
\right|
+
\frac12
\left|
\boldsymbol{\vartheta}^\top
\Bigl(
\mathbf M_{\mathcal T}(\boldsymbol{\uptheta}^\ast)
-
\mathbf M_{\mathcal S}(\boldsymbol{\uptheta}^\ast)
\Bigr)
\boldsymbol{\vartheta}
\right|.
\end{aligned}
\label{eq:hessian_split}
\end{equation}

By the curvature-control condition of EDR, for any
$\boldsymbol{\vartheta}\in B_r(\mathbf 0)$, we have:
\begin{equation}
\frac12
\left|
\boldsymbol{\vartheta}^\top
\mathbf C_{\mathcal D}(\boldsymbol{\uptheta}^\ast)
\boldsymbol{\vartheta}
\right|
\le
\varepsilon_H,
\qquad
\mathcal D\in\{\mathcal S,\mathcal T\}.
\end{equation}

Therefore,
\begin{equation}
\frac12
\left|
\boldsymbol{\vartheta}^\top
\Bigl(
\mathbf C_{\mathcal T}(\boldsymbol{\uptheta}^\ast)
-
\mathbf C_{\mathcal S}(\boldsymbol{\uptheta}^\ast)
\Bigr)
\boldsymbol{\vartheta}
\right|
\le
2\varepsilon_H.
\end{equation}

On the other hand, for each image feature
$\mathbf{z_I}$
from the source domain, assume that the corresponding image feature
$\Tilde{\mathbf{z}}_\mathbf{I}$
with the same label in the target domain satisfies:
\[
\|
\mathbf{z_I}-\Tilde{\mathbf{z}}_\mathbf{I}
\|_2
\le
\varepsilon^\prime .
\]
Since
$\mathbf M_{\mathcal D}$
is composed of cross-modal second-order coupling terms, we have:
\begin{equation}
\frac12
\left|
\boldsymbol{\vartheta}^\top
\Bigl(
\mathbf M_{\mathcal T}(\boldsymbol{\uptheta}^\ast)
-
\mathbf M_{\mathcal S}(\boldsymbol{\uptheta}^\ast)
\Bigr)
\boldsymbol{\vartheta}
\right|
\le
O(\varepsilon^\prime).
\end{equation}

Substituting the above results into Eq.~\eqref{eq:hessian_split}, we obtain:
\begin{equation}
\begin{aligned}
\frac12
\left|
\boldsymbol{\vartheta}^\top
{\mathbf H}_{\mathcal T}(\boldsymbol{\uptheta}^\ast)
\boldsymbol{\vartheta}
\right|
\le
&
\frac12
\left|
\boldsymbol{\vartheta}^\top
{\mathbf H}_{\mathcal S}(\boldsymbol{\uptheta}^\ast)
\boldsymbol{\vartheta}
\right|+
2\varepsilon_H
+
O(\varepsilon^\prime).
\end{aligned}
\end{equation}

Further substituting this into Eq.~\eqref{eq:proof_taylor} yields:
\begin{equation}
\begin{aligned}
&
\left|
{\mathcal E}_{\mathcal T}(\boldsymbol{\uptheta}^\ast+\boldsymbol{\vartheta})
-
{\mathcal E}_{\mathcal S}(\boldsymbol{\uptheta}^\ast)
\right|
\le
\left|
{\mathcal E}_{\mathcal T}(\boldsymbol{\uptheta}^\ast)
-
{\mathcal E}_{\mathcal S}(\boldsymbol{\uptheta}^\ast)
\right|\\
&\quad+\frac12
\left|
\boldsymbol{\vartheta}^\top
{\mathbf H}_{\mathcal S}(\boldsymbol{\uptheta}^\ast)
\boldsymbol{\vartheta}
\right|
+O(\varepsilon^\prime)+\frac{\rho}{6}r^3
+2\varepsilon_H.
\end{aligned}
\end{equation}

Let
$
\varepsilon
:=
O(\varepsilon^\prime)
+
\frac{\rho}{6}r^3 
$, 
then we have:
\begin{equation}
\begin{aligned}
&
\left|
{\mathcal E}_{\mathcal T}(\boldsymbol{\uptheta}^\ast+\boldsymbol{\vartheta})
-
{\mathcal E}_{\mathcal S}(\boldsymbol{\uptheta}^\ast)
\right|
\le
\left|
{\mathcal E}_{\mathcal T}(\boldsymbol{\uptheta}^\ast)
-
{\mathcal E}_{\mathcal S}(\boldsymbol{\uptheta}^\ast)
\right|
+
\frac12
\left|
\boldsymbol{\vartheta}^\top
{\mathbf H}_{\mathcal S}(\boldsymbol{\uptheta}^\ast)
\boldsymbol{\vartheta}
\right|
+
\varepsilon
+
2\varepsilon_H.
\end{aligned}
\label{eq:final_bound}
\end{equation}

Finally, since
\[
\min_{\boldsymbol{\uptheta}^\prime}
{\mathcal E}_{\mathcal T}(\boldsymbol{\uptheta}^\prime)
\le
{\mathcal E}_{\mathcal T}(\boldsymbol{\uptheta}^\ast+\boldsymbol{\vartheta}),
\]
combining this with Eq.~\eqref{eq:final_bound} gives:
\begin{equation}
\begin{aligned}
\min_{\boldsymbol{\uptheta}^\prime}
{\mathcal E}_{\mathcal T}(\boldsymbol{\uptheta}^\prime)
\le\;&
\max
\Bigl\{
{\mathcal E}_{\mathcal T}(\boldsymbol{\uptheta}^\ast),
2{\mathcal E}_{\mathcal S}(\boldsymbol{\uptheta}^\ast)
-
{\mathcal E}_{\mathcal T}(\boldsymbol{\uptheta}^\ast)
\Bigr\}
\\
&+
\frac12
\left|
\boldsymbol{\vartheta}^\top
{\mathbf H}_{\mathcal S}(\boldsymbol{\uptheta}^\ast)
\boldsymbol{\vartheta}
\right|
+
\varepsilon
+
2\varepsilon_H.
\end{aligned}
\end{equation}

Therefore, the theorem is proved.

\begin{algorithm}
   \caption{Algorithm of the proposed CRoFT}
   \label{alg:example}
\begin{algorithmic}[1]
   \STATE {\bfseries Input:} ID data $\mathbf{x^{(i)}}~(i\in{1,\cdots, N})$, ID class names of the $K$-way classification, hyperparameter $\lambda_1$ and $\lambda_2$, maximum epoch $T$.
   \FOR{$t=1$ {\bfseries to} $T$}
   \STATE Calculate the pre-trained ID image features $\mathbf{z_{I0}}$ and pre-trained language features $\mathbf{z_{T0}}$ based on the zero-shot CLIP;
   \STATE Calculate the adapted image features $\mathbf{z_I}(\mathbf{x^{(i)}},\boldsymbol{\uptheta})$ and adapted text features $\mathbf{z_T^{(j)}}(\boldsymbol{\uptheta})$;
   \STATE Generate the worst-case covariate-shifted OOD image features by Equation~\ref{eq:ood_generator};
   \STATE Minimize the classification loss of ID image features and the generated OOD image features while reshaping their energy distribution by $\widehat{\mathcal{E}}_{\mathcal{S}}(\boldsymbol{\uptheta}) + \lambda_1 \mathcal{L}_{\text{c}}+ \lambda_2 (\mathcal{L}_{\text{e}}(\mathbf{z_I}) + \mathcal{L}_{\text{e}}(\mathbf{z_I^c}))$;
   \ENDFOR
   \STATE {\bfseries Output:} Adapters' parameters $\boldsymbol{\uptheta}_{l}$.
\end{algorithmic}
\label{algoritm}
\end{algorithm}

\section{More Experiment Details}
\label{app:exp_details}
Based on the code of CoOp \cite{zhou2021learning}, we train all models with SGD optimizer with a learning rate of $2e-2$. The batch size is set to 32 except for the experiments of Tip-Adapter-F on Setup-I. Following the original paper of Tip-Adapter-F, we set the batch size as 256.
For the specific hyperparameter for each method, we follow the setting of the original paper. 

For the prompt learning methods, CoOp \cite{zhou2021learning}, CoCoOp \cite{zhou2022cocoop}, DPLCLIP \cite{APCLIP}, and Bayes-CAL \cite{zhu2023bayesian}, we use random initialization for context vectors and set the number of context tokens as 16, set the class token position (CTP) as ``end'', and set the class-specific context (CSC) as ``False''. This configuration has shown the best average performance according to CoOp’s paper.  For the DPLCLIP \cite{APCLIP} method, we set the additional hyper-parameters of DPLCLIP \cite{APCLIP} as: ``mlp\_depth=3'', ``mlp\_width=512'', and ``mlp\_dropout=0.1''.

For the CLIP-Adapter \cite{gao2023clip} method, we adopt image adapter only with the residual ratio of 0.2, and we use the bottleneck adapter with a hidden dimension that is 1/4 of the original feature dimension. This hyperparameter configuration has been demonstrated as the most effective for generic image datasets, such as ImageNet, in the original research \cite{gao2023clip}.

For the Tip-Adapter-F \cite{zhang2021tip} method, in Setup-I, 
we conduct the hyperparameter search on the learning rate and the additional hyperparameter 
of Tip-Adapter-F, i.e., $\alpha$ and $\beta$ in Tip-Adapter-F's paper. 
The corresponding hyperparameter-search results are in Table~\ref{HP-TIP-F}. According to Table~\ref{HP-TIP-F}, we select the learning rate as 0.0001, $\alpha$ as 0.5, and $\beta$ as 5.5. 
In Setup-II, we set the initial $\alpha$ as 1, while the initial beta value is searched within the range of [1, 5, 10]. This search is conducted using the validation sets to find the optimal value for $\beta$.

\begin{table}
\centering
\caption{Tip-Adapter-F's ID accuracy at different hyperparameter settings.}
\vskip 0.1in
\resizebox*{1.0\linewidth}{!}{
\begin{tabular}{llll|llllllll} 
\toprule
\multicolumn{8}{c|}{bath\_size=32, lr=0.001}                                                                                                                                                             & \multicolumn{4}{c}{{bath\_size=32, $\alpha=0.5, \beta=5.5$}}  \\ 
\cline{1-12}
$(\alpha, \beta)$ & Shot = 1   & Shot = 16  & Shot = 32  & $(\alpha, \beta)$ & Shot = 1   & Shot = 16  & \multicolumn{1}{l|}{Shot = 32}  & lr & Shot = 1     & Shot = 16    & Shot = 32                                           \\ 
\hline
(1.0, 1.5)                                        & 79.77±0.34 & 80.37±0.18 & 80.45±0.88 & (0, 5.5)                                          & 79.44±0.11 & 79.76±0.43 & \multicolumn{1}{l|}{79.85±0.47} &  &     &    &       \\
(1.0, 3.5)                                        & 80.03±0.15 & 81.11±0.60 & 80.40±0.59 & (0.5, 5.5)                                        & 79.92±0.20 & 82.40±0.88 & \multicolumn{1}{l|}{82.24±0.36} & 0.0001        & 79.45 ± 0.15 & 82.30 ± 0.93 & 83.28 ± 0.25      \\
(1.0, 5.5)                                        & 79.85±0.29 & 81.65±0.46 & 81.62±0.58 & (1, 5.5)                                          & 79.85±0.29 & 81.65±0.46 & \multicolumn{1}{l|}{81.62±0.58} & 0.0005        & 79.91 ± 0.20 & 82.18 ± 0.28 & 82.46 ± 0.30      \\
(1.0, 7.5)                                        & 79.88±0.17 & 81.57±0.14 & 82.24±0.18 & (2, 5.5)                                          & 79.95±0.17 & 80.32±0.32 & \multicolumn{1}{l|}{80.41±0.88} & 0.0010        & 79.92 ± 0.20 & 82.40 ± 0.88 & 82.24 ± 0.36      \\
(1.0, 9.5)                                        & 79.72±0.24 & 81.94±0.74 & 82.14±0.35 & (3, 5.5)                                          & 79.88±0.28 & 79.76±0.81 & \multicolumn{1}{l|}{79.57±0.55} & 0.0015        & 79.80 ± 0.07 & 81.71 ± 0.43 & 82.08 ± 0.21      \\
(1.0, 11.5)                                       & 79.65±0.36 & 81.50±0.34 & 82.38±0.41 & (4, 5.5)                                          & 79.91±0.19 & 79.51±0.63 & \multicolumn{1}{l|}{80.21±0.42} & 0.0020        & 79.44 ± 0.45 & 81.40 ± 0.31 & 82.25 ± 0.05      \\ 
\bottomrule            
\end{tabular}
}
\label{HP-TIP-F}
\end{table}

\section{More Experiment Results}
\textbf{Sensitivity analysis of hyperparameters}~
Based on 32-shot samples in Setup-I, we provide the ablation study results on the hyperparameter $\lambda_1$ and $\lambda_2$ in Table~\ref{tab:sensitivity}.
The sensitivity analysis of hyperparameters on Setup-I in our paper once again demonstrates strong evidence of CRoFT's concurrent optimization for both tasks. Our results show that the incorporation of each regularization term leads to improvements in both tasks, highlighting their effectiveness in concurrent optimization.

\begin{table}
\centering
\caption{The ablation study results on the hyperparameter $\lambda_1$ and $\lambda_2$. All results are based on 32-shot samples in Setup-I.}
\vskip0.1in
\begin{tabular}{llllll} 
\toprule
$\lambda_2$~($\lambda_1=0$) & 0     & 5     & 10    & 15    & 20    \\
\hline
AUROC                                                            & 77.25 & \textbf{85.18} & 78.56 & 80.14 & 78.41 \\
Worst-Case ACC                                                   & 42.51 & 42.84 & 42.71 & 42.91 & \textbf{43.16} \\
OOD ACC                                                           & 61.96 & 62.37 & 62.30 & 62.40 & \textbf{62.55} \\
\midrule
$\lambda_1$~($\lambda_2=0$) & 0     & 5     & 10    & 15    & 20 \\
\hline
AUROC                                                            & 77.25 & 87.07 & 87.32 & 87.38 & \textbf{87.66} \\
Worst-Case ACC                                                   & 42.51 & 43.65 & 43.71 & \textbf{43.92} & 43.61 \\
OOD ACC                                                           & 61.96 & 62.75 & 62.73 & 62.75 & \textbf{63.03} \\
\bottomrule
\end{tabular}
\label{tab:sensitivity}
\end{table}

\begin{figure*}
\centering
\centerline{\includegraphics[width=0.99\linewidth]{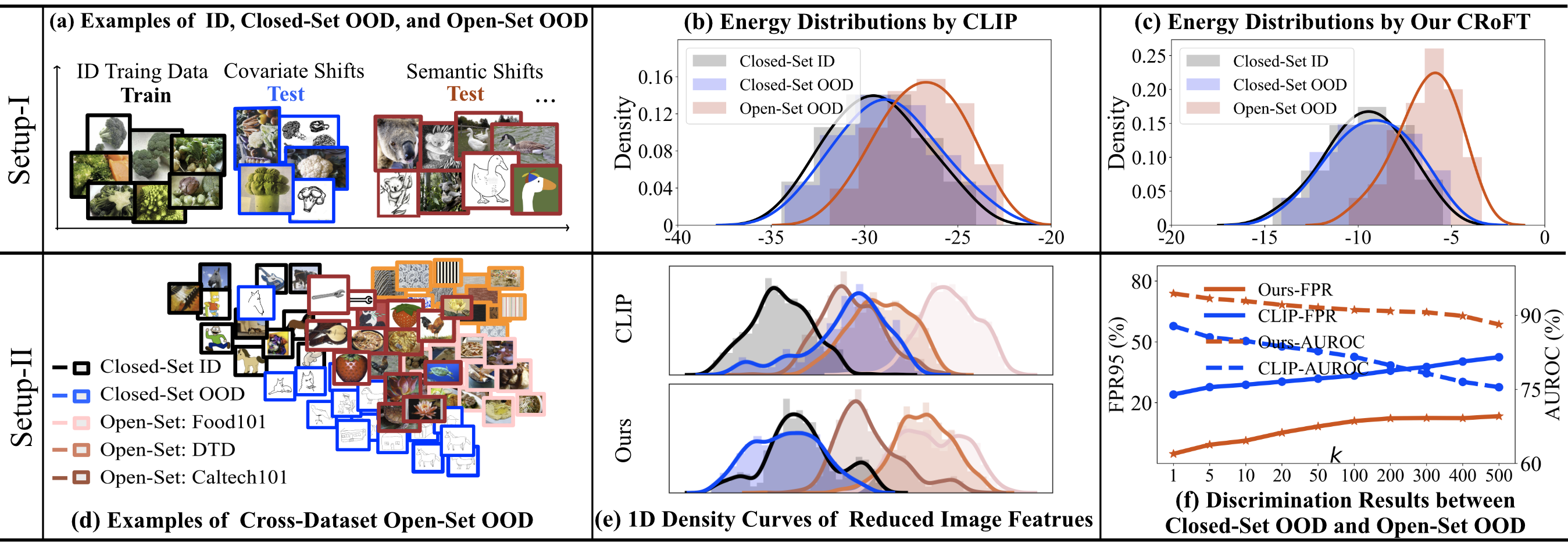}}
\caption{(a): Examples of three types of data in Setup-I: (i) closed-set ID data (e.g., broccoli), (ii) closed-set OOD data (e.g., broccoli with changed image styles), and (iii) open-set OOD data (e.g., goose). 
(b): CLIP's energy distributions on different types of data. (c): CRoFT's energy distributions on different types of data.
(d): Examples of the closed-set data and cross-dataset open-set OOD data in Setup-II, where we use the PACS dataset as the closed-set data. 
(e): Visualization of image features. All image features are reduced to the 1-dimensional space by t-SNE.
(f): The FPR95 and AUROC results in discriminating closed-set OOD and open-set OOD data.
}
\label{appfig:setups}
\end{figure*}

\textbf{Visualization on the improved open-set OOD detection}
In this section, we present more visualization results to intuitively illustrate the effectiveness of our CRoFT method in facilitating open-set OOD detection compared to the zero-shot CLIP model. As shown in Figure~\ref{appfig:setups}, CLIP's energy distributions on different types of data are highly overlapped, thus CLIP may fail to identify opens-set OOD examples. Our method significantly reduced the overlap between the energy distributions of closed-set and open-set.
For Setup-II, we visualize the distribution of image features in Figure~\ref{appfig:setups} (e), where all features are reduced to 1 dimension by t-SNE. We can observe that CLIP obtained significantly different closed-set image features, despite sharing the same categories, hindering the discrimination between the closed-set OOD and open-set OOD data. In contrast, our method can learn more discriminated image features between closed-set and open-set, improving the FPR and AUROC results when discriminating the two types of OOD data by around 20\% as shown in Figure~\ref{appfig:setups} (f).


\begin{table}
\centering
\caption{Ablation study results on $\mathcal{L}_{c}$ and $\mathcal{L}_{e}$. We report the OOD test accuracy for each domain in this table.
The best overall performances on both OOD generalization and open-set OOD detection validate the theoretical results of Theorem~\ref{Th:1} and~\ref{Th:2}.}
\vskip0.1in
\begin{tabular}{cccccccc} 
\toprule
\multirow{2}{*}{DATA} & Method  & \multicolumn{2}{c}{w~/~o $\mathcal{L}_{\text{c}}$} & \multicolumn{2}{c}{w~/~o $\mathcal{L}_{\text{e}}$} & \multicolumn{2}{c}{Our}  \\ 
\cline{2-8}
                      & V-Net   & RN50  & ViT16                & RN50  & ViT16                & RN50  & ViT16            \\ 
\hline
\multirow{7}{*}{PACS} & P       & 99.42 & 99.85                & \textbf{99.46} & 99.82                & 99.45 & \textbf{99.88}            \\
                      & A       & 94.64 & 98.35                & 94.71 & 98.68                & \textbf{94.82} & \textbf{98.94}            \\
                      & C       & 93.04 & 97.29                & 93.34 & \textbf{98.00}                & \textbf{93.38} & 97.66            \\
                      & S       & 82.08 & 91.18                & 80.56 & 90.32                & \textbf{82.16} & \textbf{92.57}            \\
                      & Avg & 92.29 & 96.67                & 92.02 & 96.71                & \textbf{92.45} &\textbf{97.26}            \\ 
\hline
\multirow{7}{*}{VLCS} & V       & \textbf{99.60} & \textbf{99.93}                & 98.37 & 99.72                & 98.94 & 99.88            \\
                      & L       & 52.99 & 54.76                & 62.52 & 55.31                & \textbf{67.09} & \textbf{61.11}            \\
                      & C       & 83.54 & 85.30                & 80.38 & 84.79                & \textbf{83.75} & \textbf{86.14}            \\
                      & S       & \textbf{73.25} & 71.80                & 69.52 & 70.41                & 68.08 & \textbf{73.73}            \\
                      & Avg & 77.34 & 77.95                & 77.70 & 77.56                & \textbf{79.47} & \textbf{80.21}            \\ 
\bottomrule
\end{tabular}
\label{tab:setup2-oodg}
\end{table}

\begin{table}
\centering
\caption{Comparison with state-of-the-art methods on few-shot benchmark datasets with CLIP RN50 pre-trained model.}
\vskip 0.1in
\resizebox*{1.0\linewidth}{!}{
\begin{tabular}{lllllllllllll} 
\toprule
Method           & OxfordPets & EuroSAT & Caltech101 & DTD   & FGVCAircraft & Flowers102 & UCF101 & Food101 & SUN397 & StanfordCars & Imagenet & Average  \\
\hline
ZS CLIP          & 85.77      & 37.56   & 86.29      & 42.32 & 17.28        & 66.14      & 61.46  & 77.31   & 58.52  & 55.74        & 60.32    & 56.69    \\
Tip,shots=1      & 86.10      & 54.38   & 87.18      & 46.22 & 19.05        & 73.12      & 62.60  & 77.42   & 61.30  & 57.54        & 60.70    & 62.33    \\
Tip,shots=2      & \textbf{87.03}      & 61.68   & 88.44      & 49.47 & 21.21        & 79.13      & 64.74  & 77.52   & 62.70  & 57.93        & 60.96    & 64.62    \\
Tip,shots=4      & 86.45      & 65.32   & 89.39      & 53.96 & 22.41        & 83.80      & 66.46  & 77.54   & 64.15  & 61.45        & 60.98    & 66.54    \\
Tip,shots=8      & 87.03      & 67.95   & 89.83      & 58.63 & 25.59        & 87.98      & 68.68  & 77.76   & 65.62  & 62.93        & 61.45    & 68.50    \\
Tip,shots=16     & 88.14      & 70.54   & 90.18      & 60.93 & 29.76        & 89.89      & 70.58  & 77.83   & 66.85  & 66.77        & 62.01    & 70.32    \\
\hline
Tip-F,shots=1    & \textbf{87.00}      & 59.53   & 89.33      & 49.65 & 20.22        & 79.98      & 64.87  & 77.51   & 62.50  & 58.86        & 61.13    & 64.60    \\
Tip-F,shots=2    & \textbf{87.03}      & 66.15   & 89.74      & 53.72 & 23.19        & 82.30      & 66.43  & 77.81   & 63.64  & 61.50        & 61.69    & 66.65    \\
Tip-F,shots=4    & 87.54      & 74.12   & 90.56      & 57.39 & 25.80        & 88.83      & 70.55  & 78.24   & 66.21  & 64.57        & 62.52    & 69.67    \\
Tip-F,shots=8    & 88.09      & 77.93   & 91.44      & 62.71 & 30.21        & 91.51      & 74.25  & 78.64   & 68.87  & 69.25        & \textbf{64.00}    & 72.45    \\
Tip-F,shots=16   & 89.70      & \textbf{84.54}   & 92.86      & 66.55 & 35.55        & 94.80      & 78.03  & 79.43   & 71.47  & 75.74        & \textbf{65.51}    & 75.83    \\
\hline
Adapter,shots=1  & 85.99      & 61.40   & 88.60      & 45.80 & 17.49        & 73.49      & 62.20  & 76.82   & 61.30  & 55.13        & 61.20    & 62.67    \\
Adapter,shots=2  & 86.73      & 63.90   & 89.37      & 51.48 & 20.10        & 81.61      & 67.12  & 77.22   & 63.29  & 58.74        & 61.52    & 65.55    \\
Adapter,shots=4  & 87.46      & 73.38   & 89.98      & 56.86 & 22.59        & 87.17      & 69.05  & 77.92   & 65.96  & 62.45        & 61.84    & 68.61    \\
Adapter,shots=8  & 87.65      & 77.93   & 91.40      & 61.00 & 26.25        & 91.72      & 73.30  & 78.04   & 67.50  & 67.89        & 62.68    & 71.40    \\
Adapter,shots=16 & 87.84      & 84.43   & 92.49      & 65.96 & 32.10        & 93.90      & 76.76  & 78.25   & 69.55  & 74.01        & 63.59    & 74.44    \\
\hline
CoOp,shots=1     & 85.89      & 50.63   & 87.53      & 44.39 & 9.64         & 68.12      & 61.92  & 74.32   & 60.29  & 55.59        & 57.15    & 59.59    \\
CoOp,shots=2     & 82.64      & 61.50   & 87.93      & 45.15 & 18.68        & 77.51      & 64.09  & 72.49   & 59.48  & 58.28        & 57.81    & 62.32    \\
CoOp,shots=4     & 86.70      & 70.18   & 89.55      & 53.49 & 21.87        & 86.20      & 67.03  & 73.33   & 63.47  & 62.62        & 59.99    & 66.77    \\
CoOp,shots=8     & 85.32      & 76.73   & 90.21      & 59.97 & 26.13        & 91.18      & 71.94  & 71.82   & 65.52  & 68.43        & 61.56    & 69.89    \\
CoOp,shots=16    & 87.01      & 83.53   & 91.83      & 63.58 & 31.26        & 94.51      & 75.71  & 74.67   & 69.26  & 73.36        & 62.95    & 73.42    \\
\hline
\textbf{Ours,shots=1}  & {86.43} & \textbf{64.64} & \textbf{89.49} & \textbf{50.18} & \textbf{21.81} & \textbf{81.45} & \textbf{66.30} & \textbf{77.36} & \textbf{63.03} & \textbf{58.86} & \textbf{61.61} & \textbf{65.56}  \\
\textbf{Ours,shots=2}  & {86.86} & \textbf{67.60} & \textbf{90.47} & \textbf{55.32} & \textbf{24.51} & \textbf{85.91} & \textbf{69.07} & \textbf{77.90} & \textbf{65.87} & \textbf{62.17} & \textbf{62.31} & \textbf{68.00}  \\
\textbf{Ours,shots=4}  & \textbf{88.28} & \textbf{75.15} & \textbf{91.44} & \textbf{59.34} & \textbf{27.06} & \textbf{91.47} & \textbf{71.72} & \textbf{78.35} & \textbf{68.05} & \textbf{66.61} & \textbf{62.88} & \textbf{70.94}  \\
\textbf{Ours,shots=8}  & \textbf{88.96} & \textbf{78.23} & \textbf{92.17} & \textbf{63.24} & \textbf{33.39} & \textbf{94.03} & \textbf{75.50} & \textbf{78.36} & \textbf{69.79} & \textbf{70.17} & {63.78} & \textbf{73.42}  \\
\textbf{Ours,shots=16} & \textbf{89.97} & \textbf{84.54} & \textbf{93.27} & \textbf{67.26} & \textbf{38.85} & \textbf{95.66} & \textbf{78.43} & \textbf{79.44} & \textbf{71.90} & \textbf{76.09} & {65.20} & \textbf{76.42}  \\
\bottomrule
\end{tabular}
}
\label{tab:11-res}
\end{table}

\begin{table}
\centering
\caption{Energy score for the closed-set OOD, closed-set ID, generated closed-set OOD, and open-set OOD data. Based on Setup-I and our 32-shot fine-tuned CRoFT model, we present the 5th, 25th, 50th, 75th, and 95th percentiles of energy scores.}
\vskip 0.1in
\begin{tabular}{llllll} 
\toprule
DATA                     & 0.05   & 0.25   & 0.50   & 0.75   & 0.95  \\
\hline
ID                       & -17.36 & -13.59 & -12.08 & -10.04 & -8.00 \\
Closed-Set OOD           & -16.90 & -13.34 & -11.77 & -9.79  & -7.75 \\
Generated Closed-Set OOD & -17.19 & -13.42 & -11.89 & -9.84  & -7.78 \\
Open-Set OOD             & -12.76 & -9.99  & -8.55  & -6.88  & -4.97 
\\
\bottomrule
\end{tabular}
\label{tab:distribution_energy}
\end{table}

\textbf{CRoFT's effectiveness on robust few-shot learning}
We conducted additional experiments on benchmark datasets commonly used for CLIP-based methods, including Calteh101, Oxford Flowers, DTD, FGVCAircraft, UCF101, Oxford Pets, and EuroSAT. Based on CLIP RN50, the 1-shot, 2-shot, 4-shot, and 16-shot results of our method and the SOTA method Tip-Adapetr-F are reported in Table~\ref{tab:11-res}. The results show that our method achieves higher test accuracy in most cases and obtains the best average results under various training shots. CRoFT also outperforms Tip-Adapetr-F stably. It is evident that CRoFT also shows stable improvements under very-low shot conditions, with about a 1.5\% boost on average under the 1-shot and 2-shot settings.

\textbf{Energy distribution of the generated worst-case covariate-shifted OOD features}
We present the energy distribution of the generated worst-case covariate-shifted OOD features to highlight the quality of these OOD features. Based on Setup-I and our 32-shot fine-tuned CRoFT model, we present the 5th, 25th, 50th, 75th, and 95th percentiles of energy scores for the generated closed-set OOD, closed-set ID, closed-set OOD, and open-set OOD across 3 runs. 
As shown in Table~\ref{tab:distribution_energy}, our experiment results demonstrate that the energy distribution of the generated worst-case covariate-shifted OOD feature closely resembles that of the closed-set OOD data. By optimizing these high-quality generated OOD features, we can enhance the model's generalization capacity for closed-set OOD data, thereby improving the model's ability to discriminate between closed-set OOD data and open-set OOD data.

\end{document}